\documentclass[11pt]{article}

\usepackage[a4paper,
            left=1in,
            right=1in,
            top=1in,
            bottom=1in,
            ]{geometry}

\usepackage{natbib}
 \setcitestyle{aysep={}}	
 \renewcommand{\cite}{\citep}	

\usepackage{hyperref}
\usepackage{xcolor}
\usepackage{subcaption}   
\usepackage{booktabs}
\usepackage{tabularx}
\usepackage{makecell}
\usepackage{amsmath,amssymb}
\usepackage{soul}
\usepackage{arydshln}
\usepackage[size=tiny,disable]{todonotes}
\usepackage{soul}
\usepackage{tcolorbox}
\usepackage{arydshln}
\definecolor{darkblue}{rgb}{0, 0, 0.5}
\hypersetup{colorlinks=true,citecolor=darkblue, linkcolor=darkblue, urlcolor=darkblue}

\newcommand{\limit}[1]{\begin{tcolorbox}\textbf{Limitations}: #1\end{tcolorbox}}

\newcommand{\wei}[1]{\textcolor{black}{#1}}

\newcommand{\CL}[1]{\todo[color=green!40]{\textit{CL:} #1}}

\newcommand{\attacked}[1]{\textcolor{orange}{#1}}

\begin{document}


\title{Towards Explainable 
Evaluation Metrics for \\ 
Natural Language Generation
}

\author{Christoph Leiter$^1$, Piyawat Lertvittayakumjorn$^2$, Marina Fomicheva$^3$, \\ Wei Zhao$^4$, Yang Gao$^5$, Steffen Eger$^1$ \\ 
$^1${\small Technische Universität Darmstadt}, 
$^2${\small Imperial College London}, $^3${\small University of Sheffield}\\
$^4${\small Heidelberg Institute for Theoretical Studies}, $^5${\small Google Research}}
\date{}










\maketitle

\begin{abstract}
Unlike classical lexical overlap metrics such as BLEU, most current evaluation metrics (such as BERTScore or MoverScore)
are based on black-box language models such as BERT or XLM-R.  
They often achieve strong correlations with human judgments, but recent research indicates that the lower-quality 
classical 
metrics remain dominant, 
one of the potential reasons being that their decision processes are  
transparent. 
To foster more widespread acceptance of the novel high-quality metrics, explainability thus becomes crucial. In this concept paper, we identify key properties and  
propose key goals 
of \emph{explainable machine translation evaluation metrics}. 
We also  
provide a synthesizing overview over recent approaches for explainable machine translation metrics and discuss how they 
relate to those goals and properties. Further, we  conduct own novel experiments, which (among others) find that current 
adversarial NLP techniques are unsuitable for automatically identifying limitations of high-quality black-box evaluation metrics, as they are not meaning-preserving. Finally, we provide a vision of future approaches to explainable evaluation metrics and their evaluation.  
We hope that our work can help catalyze and guide future research on explainable evaluation metrics and,  mediately, also contribute to better and more transparent text generation systems. 
\end{abstract}




\section{Introduction} \label{sec:intro}
\begin{figure}[!htb]
    \includegraphics[scale=0.7]{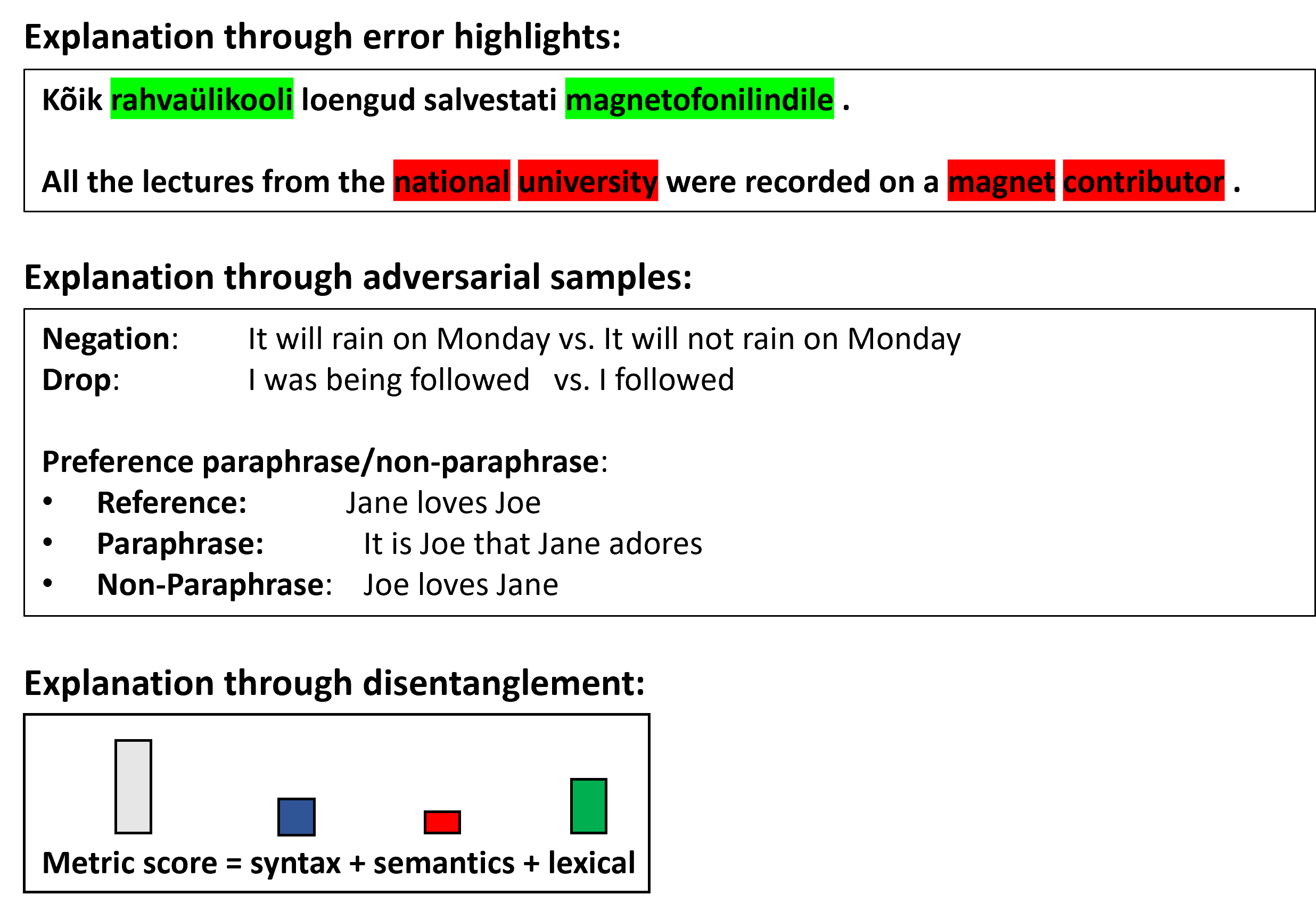}
    \caption{
    Current explanations for NLG evaluation metrics fall into three categories.
    \emph{Explanation through error highlights} explains a metric score by highlighting word-level errors in a translation. \emph{Explanation through adversarial samples} checks whether the metrics perform reasonably on adversarial samples. \emph{Explanation through disentanglement} splits the metric score into different components.}
    \label{fig:illustration}
\end{figure}

{
The field of evaluation metrics for Natural Language Generation (NLG) is currently in a deep crisis: 
While multiple high-quality evaluation metrics \citep{zhao-2019,zhang-2020,rei-2020,sellam-2020,yuan-2021}
have been developed in the last few years, 
the Natural Language Processing (NLP) community seems reluctant to adopt 
them to assess NLG systems \citep{marie-etal-2021-scientific,gehrmann2022repairing}. 
In fact, the empirical investigation of \citet{marie-etal-2021-scientific} shows that the vast majority of machine translation (MT) papers 
(exclusively) relies on surface-level evaluation metrics like BLEU and ROUGE \citep{papineni-2002,lin-2004} for evaluation, which were invented two decades ago, and the situation has allegedly even worsened recently.  
These surface-level metrics cannot measure semantic similarity of their inputs and are thus fundamentally flawed, particularly when it comes to assessing the quality of recent state-of-the-art NLG systems \citep{peyrard-2019-studying}, calling the credibility of a whole scientific field in question. 

We argue that the potential 
reasons for this 
neglect of recent high-quality metrics include:
(i) non-enforcement by reviewers; (ii) easier comparison to previous research, e.g., by copying BLEU-based results from tables of related work  (potentially a pitfall in itself); (iii) 
computational inefficiency to run expensive new metrics at large scale; 
(iv) lack of trust in and transparency of high-quality black box metrics.

In this work, we concern ourselves with the last named reason, 
\emph{explainability}. 
In recent years, explainability in Artificial Intelligence
(AI) has been developed and studied extensively due to several needs \cite{Samek2018Overview,vaughan2020human}.
For users of the AI systems, explanations help them make more informed decisions
(especially in high-stake domains) 
\cite{sachan2020explainable,lertvittayakumjorn-etal-2021-supporting}, 
better understand and hence gain trust of the AI systems
\cite{pu2006trust,toreini2020relationship},
and even learn from the AI systems to accomplish the tasks more successfully
\cite{mac2018teaching,lai2020chicago}.
For AI system designers and developers, explanations allow them
identify the problems and weaknesses of the system
\cite{krause2016interacting,han-etal-2020-explaining},
calibrate the confidence of the system \cite{zhang2020effect},
and improve the system accordingly \cite{kulesza2015principles,lertvittayakumjorn-2021}.
%
%
%
%

Explanability is particularly desirable for evaluation metrics.
%
\citet{sai-2020} 
suggest that explainable NLG metrics should focus on providing more information than 
just a single score (such as fluency or adequacy). 
\citet{celikyilmaz-2020} stress the need for explainable evaluation metrics 
to spot system quality issues and to achieve a higher trust in the evaluation of NLG systems. 
%
Explanations indeed play a vital role in building trust for new evaluation metrics.\footnote{As one illustrating example, we mention the recent paper of \citet{moosavi2021scigen} (also personal communication with the authors). In the paper, the authors express their distrust for the metric BERTScore applied to a novel text generation task. In particular, they point out that the BERTScore of a non-sensical output is 0.74, which could be taken as an unreasonably high value. While BERTScore may indeed be unsuitable for their novel task, the score of 0.74 is meaningless here, as evaluation metrics may have arbitrary ranges. In fact, BERTScore typically has a particularly narrow range, so a score of 0.74 even for bad output may not be surprising. Explainability techniques would be very helpful in preventing such misunderstandings.}
For instance, if the explanations for the scores 
align well with human reasoning, 
the metric will 
likely be better accepted by 
the research community.
By contrast, if the explanations are counter-intuitive, 
users and developers will lower their trust and be 
alerted to take additional 
actions, 
such as trying to improve the metrics using insights from the explanations 
or 
looking for alternative metrics that are more trustworthy. 
Furthermore, 
explainable metrics can used for other purposes:
e.g., when a metric produces a low score for a given input,
the highlighted words (a widely used method
for explanation, see Section \ref{sec:taxonomy}) in the input are
natural candidates 
for 
manual post-editing. 
%


This concept paper aims at providing a systematic overview over the existing efforts
in explainable NLG evaluation metrics and at an outlook for promising future research directions.
We first provide backgrounds on evaluation metrics (Section \ref{sec:metrics}) 
and explainability (Section \ref{sec:xai}), and discuss the goals and properties
of explainable evaluation metrics (Section \ref{sec:xmte}). 
Then we review and discuss existing datasets (Section \ref{sec:datasets}) 
and methods (Section \ref{sec:taxonomy})
for explainable metrics, covering both \emph{local} and \emph{global} 
methods for providing explanations; see Figure \ref{fig:illustration} for an overview. 
%
To better understand the properties (e.g. faithfulness and 
plausibility) of different
explainable evaluation metrics, we test and compare
multiple representative methods under three
newly-proposed experimental setups and present
the results in Section \ref{sec:new}. 
Our focus in this context is particularly on the relation between explainability and adversarial techniques, where we examine whether the latter can automatically identify limitations of existing evaluation metrics, thereby promoting a better understanding of their weaknesses. 
At last, we discuss promising ideas for future work
(Section \ref{sec:future}) and conclude the paper (Section \ref{sec:conclusion}).  
}
%

We mainly focus on the explainable evaluation 
metrics for MT in this work,
as it is one of the most representative NLG tasks. 
Most of our observations and discussions 
can easily be adapted to other NLG tasks, however.
Source code for our experiments are publicly available at \url{https://github.com/Gringham/explainable-metrics-machine-translation}.


\section{Evaluation Metrics} \label{sec:metrics}
\subsection{Metric Dimensions}

We first differentiate evaluation metrics along several dimensions. 
A summary is shown in Table \ref{table:metrics}.\footnote{We note that categorizations of metrics along many different dimensions are possible.}
In the following, we will use the abbreviation MTE to stand for `Machine Translation Evaluation' (metrics). 

\begin{table}[!htb]
\centering
    \begin{tabularx}{\textwidth}{lXX}
     \toprule
     \textbf{Dimension} &  \textbf{Description} \\ 
     \midrule
     \emph{Type of input} & Whether source, reference translation or both are used as benchmark for comparison \\ \hline
     \emph{Input representation} & Whether the metric relies on words or uses a continuous representation of the input such as word embeddings \\ \hline 
     \emph{Supervision} & 
     The degree of supervision a metric uses \\ 
     \hline
     \emph{Granularity} & At which level a metric operates: Word-level, sentence-level, document-level \\ \hline
     \emph{Quality aspect} & What a metric measures: Adequacy, fluency, etc. \\ \hline
     \emph{Learning objective} & How a metric is induced: Regression, ranking, etc. \\ \hline
     \emph{Interpretability} & Whether the metric  
     is interpretable \\
     \bottomrule
\end{tabularx}
\caption{A typology of categorizations for evaluation metrics.}
\label{table:metrics}
\end{table}

\paragraph{Type of input} The majority of metrics for MT compare a human reference to an MT hypothesis \citep{zhang-2020,zhao-2019,colombo-etal-2021-automatic,sellam-2020}. Some metrics directly compare source texts to MT hypotheses instead \citep{zhao-2020,song-etal-2021-sentsim,lo-larkin-2020-machine}, which is a more difficult task, but eliminates the need for an expert reference. In contrast, there are also metrics that leverage all three  information signals, i.e., source, reference, and MT hypothesis \citep{rei-2020}, which may lead to better performances when they are all available. 

\paragraph{Input representation}
Embeddings have seen great success in encoding linguistic information as dense vector representations \citep{mikolov-2013,devlin-2019}. The following distinctions of embedding usage can be found 
in 
\citet{sai-2020}:
\begin{itemize}
\item\textbf{No Embeddings:}
Metrics such as BLEU \citep{papineni-2002} or NIST \citep{doddington-2002} are based on exact word matches between n-grams of hypothesis and reference sentences. METEOR \citep{lavie-2004,banerjee-2005} goes a step further and employs 
a.o.\ stemming 
to account for non-exact word matches. Nonetheless, all of these metrics do not check whether the meaning of two sentences is preserved and have difficulties to deal with paraphrases and 
synonyms 
\citep{ananthakrishnan-2006, reiter-2018}. These problems lead to a lower correlation with human judgements in segment-level evaluations 
\citep{stanojevic-2015, mathur-2020}. 

\item\textbf{Static Embeddings:} 
Static embeddings are vector representations 
such that the representation of a token will 
be the same across contexts \citep{mikolov-2013,pennington-2014}. 
The use 
of static embeddings in MTE allows to compare sentences that use different words than a reference but express the same content \citep{fishel-2017}. Examples of metrics based on static word embeddings are 
bleu2vec \citep{fishel-2017} and 
the approach of \citet{ng-abrecht-2015-better}, who induce metrics for summarization.  

\item\textbf{Contextualized Embeddings:}
Contextualized embeddings assign vector representations depending on an input context \citep{wang-2020,devlin-2019}. 
Comparison between sentences in MTE that build on contextualized representations can be at word-level \citep{zhang-2020, zhao-2019} or by comparing embeddings of the whole sentence \citep{reimers-2020}. The currently best-performing metrics build on contextualized representations, e.g.\ MoverScore \citep{zhao-2019}, BERTScore \citep{zhang-2020} and BaryScore \citep{colombo-etal-2021-automatic}. 

\end{itemize}
As metrics without embeddings perform a strict comparison between words, we will refer to them as \textit{hard} metrics \citep{zhao-2020}. In contrast, we will refer to metrics that use embeddings as \textit{soft} metrics.

\paragraph{Supervision}
This dimension distinguishes whether an automated MTE metric 
uses a form of supervision or not. 
\begin{itemize}
\item\textbf{Reference-Based:} We call a metric 
reference-based if it requires one or multiple human reference translations to compare with the hypothesis, which can be seen as a form of supervision.  
 Such a metric cannot be used in cases where no bilingual expert 
is available. 

\item\textbf{Reference-Free:} 
Reference-free metrics do not require a reference translation to grade the hypothesis \citep{zhao-2020,ranasinghe-2020,song-etal-2021-sentsim}. 
Instead, they 
directly compare the source to the hypothesis. 
In the literature, reference-free MTE is also referred to as ``reference-less'' or (especially in the MT context) ``quality estimation''. 

\end{itemize}

Another distinction is whether a metric is untrained or trained, i.e.\ whether it has trainable weights that are fitted in order to solve the task of MTE \citep{celikyilmaz-2020}:
\begin{itemize}
\item\textbf{Not Trainable:} This dimension indicates that a metric has no learnable weights. 
\item\textbf{Pre-Trained:} Some metrics leverage embeddings that were trained in more general tasks without fitting them to the task of MTE. This is the case for BERTScore \citep{zhang-2020} and MoverScore \citep{zhao-2019} which extract contextualized embeddings from language models.

\item\textbf{Fine-Tuned:} Finally, some metrics directly optimize the correlation of metric scores with human judgment in a supervised learning setup. 
Examples of this type of metrics include BLEURT \citep{sellam-2020}, COMET \citep{rei-2020}, and TransQuest \citep{ranasinghe-2020}. 
\end{itemize}

\citet{Belouadi2022USCOREAE} argue for metrics that use no form of supervision in order to be as inclusive as possible, i.e., applicable in settings where no supervision signals are available (e.g., for low-resource and non-English language pairs). 

\paragraph{Granularity}{
Translation quality can be evaluated at different levels of granularity: word-level, sentence-level and document-level.
The majority of metrics for MT give a single sentence-level score to each input. Beyond individual sentences, a metric may also score whole documents (multiple sentences) \citep{zhao2022discoscore,jiang2021blond}, an avenue that is likely to dominate in the future as text generation systems at document-level will become more common. In the MT community, metrics that evaluate translations at the word-level (e.g., whether individual words are correct or not) are also common \citep{Turchi2014AdaptiveQE,shenoy2021investigating}.\todo{SE: @MF?} 
} 

\paragraph{Quality aspect} This refers to what a metric evaluates and measures (or is at least compared to), e.g., human assigned adequacy (via direct assessment \citep{graham-2016} obtained from crowd-workers), fluency, or other aspects (such as relevance, informativeness, or correctness, mostly in other NLG fields). 
Recently, there has been a tendency to compare metrics 
to human scores derived from fine-grained error analysis \citep{freitag-2021} as they seemingly correspond better to human professional translators. 
\todo{SE: HTER in this aspect also? @MF, @CL? Does it make sense what I wrote? \\CL: To me it makes sense, yes. I think I wouldn't mention the HTER at this point}

\paragraph{Learning objective}
%
\citet{yuan-2021} describe a further distinction based on the task a metric is designed to solve (or \emph{how} a metric is implemented) in order to achieve high correlation with human judgments. They identify the three tasks below. 
\begin{itemize}
    \item \textbf{Unsupervised Matching:} They place all metrics that match hypothesis and reference tokens in this dimension. In terms of the dimension ``untrained vs. trained'',  unsupervised matching captures metrics that are ``not-trainable'' or ``pre-trained''.
    \item \textbf{Supervised Regression/Ranking:} These models use supervised regression to train a metric to predict continuous values that directly correlate with human judgments from hypothesis and reference tokens.
    Alternatively, a metric may be trained to rank hypothesis sentences. For example, \citet{rei-2020} propose a 
    ranking setup in which they minimize a loss that becomes smaller the greater the distance between embedding representations of the two sentences become.  
    \item \textbf{Text Generation:} 
    The authors place the metric PRISM \citep{thompson-2020} and their own metric BARTScore in this condition. BARTScore leverages the fact that a model that is trained to generate a hypothesis from a reference or source is at the same time trained to maximize the correlation to human judgement. 
    They calculate the generation probability of the ground truth given the input. 
    As metric, they take the arithmetic mean of the loss scores of the directions reference $\rightarrow$ hypothesis and hypothesis $\rightarrow$ reference.
\end{itemize}

\paragraph{Interpretability} There is a general tradeoff between quality and interpretability of evaluation metrics: metrics that follow simple formulas such as BLEU and ROUGE are easily interpretable but of low quality, while high-quality metrics based on sophisticated language models are typically black box. Aiming for metrics that satisfy both is the main concern of our paper.

\section{Explainability and Explanations}
\label{sec:xai}
 
\begin{figure}[!tbp]
    \centering
    \includegraphics[width=0.45\textwidth]{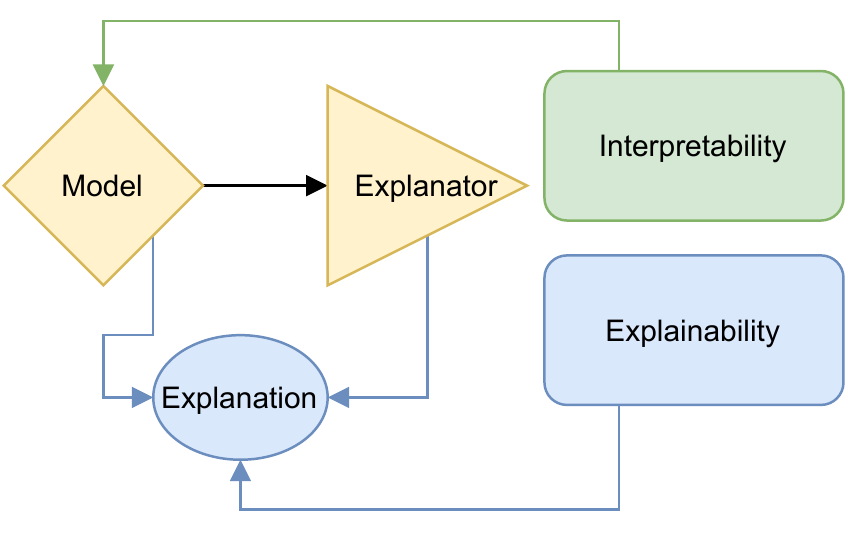}
  \caption{An illustration of passive interpretability versus active explainability.}
  \label{fig:InterprExpl}
\end{figure}

In this section, we introduce 
core concepts 
of explainable artificial intelligence (XAI)
with a focus on their use in NLP.
Generally, XAI aims to systematically expose complex AI models to humans in 
a comprehensible
manner \cite{samek2019explainable}.
This amounts to \emph{interpreting} and \emph{explaining} the models, and prior work has noted that there is an ambiguous usage of the terms \textit{interpretability} and \textit{explainability} \citep{barredo-arrieta-2020, jacovi-2020}. In this paper, we follow the notion of \emph{passive interpretability} and \emph{active explainability} \citep[e.g.][]{barredo-arrieta-2020, bodria-2021}, as illustrated in Figure \ref{fig:InterprExpl}. A model is explainable if the model itself or some external instance (described as \textit{Explanator} in the figure) can actively provide explanations that make the decision process or the output clearer to a certain audience. In contrast, a model is interpretable if it can be passively understood by humans (e.g., decision trees and k-nearest neighbours).

As discussed in Section~\ref{sec:metrics}, modern MTE metrics 
are mostly based on embeddings and black-box language models, making the metrics become non-interpretable.
Recent work, therefore, tries to generate explanations for the metrics to help humans better understand them \citep[e.g.][]{freitag-2021,fomicheva-2021b}.
Hence, the remainder of this section will discuss several aspects of \textit{explanations} in XAI and NLP literature, including their scopes, forms, sources, and evaluations, to serve as an essential background for the rest of the paper.  

\subsection{Scopes of Explanations} \label{subsec:scopeexp}
The scope of an explanation indicates the subject that the explanation explains. 
Existing surveys \citep[e.g.,][]{doshi-velez-2017,arya-2019,danilevsky-2020,bodria-2021}
classify explanations mainly into two scopes -- local and global explanations.
\begin{itemize}
    \item \textbf{Local explanations} aim to explain a particular output of the model.
    Specifically, given a model $M$ and an input $x$, a local explanation explains why the model outputs $M(x)$ for the input $x$. Examples of local explanation methods are \citet{bach-2015,ribeiro-2016,koh2017understanding,wallace-etal-2018-interpreting}.
    \item \textbf{Global explanations} aim to explain the working mechanism of the model $M$ in general, regardless of a specific input. 
    To do so, it may require a set of (real or synthetic) examples in order to observe and summarize the model behaviour to be a global explanation. 
    Examples of global explanation methods are \citet{sushil-etal-2018-rule,tan2018distill,lundberg2020local,setzu2021glocalx}.
\end{itemize}

Besides, \citet{lertvittayakumjorn-2021} recognize that there are explanations which stay between the local and the global scopes.
These amount to explanations for groups of examples such as a group of false positives of a certain class and a cluster of examples with some fixed features. Examples of XAI work that fall into this middle scope are \citet{chan2020subplex,johnson2020njm}.

\subsection{Forms of Explanations}
Explanations for AI models and their predictions can be presented in various forms. 
Concerning \textbf{local explanations}, some methods \citep[e.g.,][]{lei-etal-2016-rationalizing,jain-etal-2020-learning} identify or extract parts in the input $x$ that $M$ considers important when predicting $M(x)$. The extracted parts can be called the \textit{rationales} or \textit{highlights} for $M(x)$.
Some methods, instead, assign \textit{relevance scores} (also called \textit{importance scores} or \textit{attribution scores}) to all the features (usually tokens) in $x$, showing their relative importance towards $M(x)$ \citep[e.g.,][]{ribeiro-2016,sundararajan-2017,kim-2020}. These scores could be visualized using a saliency map laid on the input $x$.
Another way to relate the input to the predicted output is using \textit{decision rules} as explanations, showing the sufficient conditions (satisfied by $x$) for $M$ to predict $M(x)$.
For example, a review $x$ is classified as a positive review by $M$ because of the decision rule $r:$ \texttt{If x contains ``best'' \& ``enjoy'', then M(x) = Positive} with the confidence of 92\%.
Exemplary Methods using decision rule explanations are \citet{ribeiro-2018} and \citet{tang-surdeanu-2021-interpretability}.
Besides, $M(x)$ could also be explained using synthetic or natural input examples.
For example, by using influence functions, \citet{koh2017understanding} present training examples that are most responsible for making $M$ predict $M(x)$ as explanations.
\citet{wachter-2018} explains $M(x)$ using a \textit{counterfactual example} -- an (optimized) input $x'$ that is closest to $x$ but has a different prediction (i.e., $M(x') \neq M(x)$). 
Last but not least, there are also methods producing \textit{natural language texts} as explanations for $M(x)$ such as the work by \citet{camburu2018snli} and \citet{liu-etal-2019-towards-explainable}.

To explain the global behavior of the model $M$, in contrast, there are some applicable forms of \textbf{global explanations}.
One is to train an interpretable model $S$ to emulate (the behaviour of) the model $M$ and then use $S$ as a global explanation of $M$. We call $S$ a \textit{surrogate model} \cite{sushil-etal-2018-rule,tan2018distill,lakkaraju2019faithful}.
Otherwise, we may understand the model $M$ by observing a collection of its local explanations that are diverse enough to cover most regions in the input space  \cite{ribeiro-2016,ribeiro-2018}.
Some existing work further aggregates such local explanations to be a single global explanation of $M$ \citep[e.g.,][]{lundberg2020local,setzu2021glocalx}. 
Additionally, to facilitate human understanding, one may also incorporate high-level concepts associated to the input (e.g., existence of semantic concepts for image classification \citep{kim2018interpretability} and syntactic relationships between two input sentences for sentence-pair classification \citep{mccoy-etal-2019-right}) into the explanation.
These methods could be categorized as \textit{concept attribution} methods \citep{bodria-2021}.

Note, however, that 
besides the 
global explanations discussed above,
there are also several \textbf{analysis methods} devised to further understand $M$, e.g., analyzing behaviours of individual neurons \cite{li-etal-2016-visualizing,poerner-etal-2018-interpretable,mu2020compositional}, probing knowledge stored in the model \cite{hewitt-manning-2019-structural,voita-titov-2020-information,belinkov2021probing}, testing model behaviour in many (challenging) scenarios \cite{poliak-etal-2018-collecting,wang2019superglue,ribeiro-etal-2020-beyond}, searching for adversarial rules where the model often fails \cite{ribeiro-etal-2018-semantically,wallace-etal-2019-universal}, etc.
Though it is arguable whether these analysis methods are producing \textit{explanations} for the model, they do provide lots of useful global insights into the model. 
So, we group them under global techniques to be discussed, specifically for MTE, in Section \ref{sec:global_techniques}. 

\subsection{Sources of Explanations}
In a specific situation, ways to obtain explanations depend largely on the category of the model $M$. Here, we list the three main categories.
\begin{itemize}
    \item \textbf{Interpretable models.} Some machine learning models are easy to understand for humans such as Naive Bayes, decision trees, linear regression, and k-nearest neighbours \cite{freitas2014comprehensible}. Existing works, therefore, call them \textit{interpretable} \cite{arya2019one} or \textit{self-explaining} \cite{danilevsky-2020}. 
    We can normally obtain local explanations of these models at the same time as predictions (e.g., the corresponding path in the decision tree), while the models can be considered global explanations of themselves (e.g., the trained decision tree itself). 
    
    \item \textbf{Explainable models by design.} Some models are not interpretable, still they are able to output local explanations for their predictions by design.
    For instance, \citet{narang2020wt5} train a black-box sequence-to-sequence model to perform a set of tasks and also output corresponding explanations if requested. 
    \citet{jain-etal-2020-learning} train a black-box model to extract a rationale from a given input first and then feed the rationale into another model to predict the output. So, the extracted rationale can be used as a local explanation for the prediction.
    However, because these models are not interpretable, we cannot use themselves as global explanations.
    Therefore, we require a post-hoc global explanation method in this case.
    
    \item \textbf{Black-box models.} These models are not interpretable and do not produce explanations for their predictions. Therefore, they are completely black-boxes and require a separate step to extract (local or global) explanations from them in particular. 
    Methods for extracting explanations in this case are called \textit{post-hoc explanation methods}. 
    Some of them can be applied to any models (i.e., being \textit{model-agnostic}) such as SHAP \cite{lundberg-2017} and LIME \cite{ribeiro-2016}, whereas others are applicable to a specific class of models (i.e., being \textit{model-specific}) such as Tree SHAP \cite{lundberg2018consistent} for tree-based models and LRP \cite{bach-2015} for neural networks.
\end{itemize}

Note that, although post-hoc explanation methods are often used with black-box models, they can be applied to models from the first two categories to generate alternative explanations.

\subsection{Evaluation of Explanations} 
\label{EvalOfExplanations}
In general, there are numerous methods to evaluate explanations in XAI literature.
\citet{pljthesis} divides them into two broad categories, i.e., \textit{intrinsic evaluation} and \textit{extrinsic evaluation}.
While the former is concerned with evaluating desirable properties of explanations, the latter is concerned with evaluating the usefulness of the explanations in a downstream task, which usually requires a combination of several intrinsic properties. 
In this section, we will introduce two intrinsic properties of explanations, \textit{faithfulness} and \textit{plausibility}, together with existing methods for evaluating them, 
  seeing that these two properties are relevant to the context of explainable MTE metrics. 

\paragraph{Faithfulness} \label{Faithfulness}
An explanation is faithful if it accurately represents the reasoning process underlying the model’s prediction(s) \cite{jacovi-2020}.
In other words, faithfulness focuses on helping humans understand how the model $M$ works.
Existing faithfulness evaluation methods could be categorized into three groups with regards to the \textit{ground truths} that we compare the explanations against. 

\begin{itemize}
    \item \textbf{Known ground truths.} 
    For an interpretable model, we can treat the model's decision making mechanism, which are transparent to humans, as a ground truth for evaluating faithfulness of an alternative explanation.
    For instance, if we explain a prediction of a linear regression model, the attribution score of a feature should be consistent with the product of the feature and the associated weight in the model. 
    This could be seen as a sanity check for faithfulness of the explanation method.
    
    \item \textbf{Inferred ground truths.} 
    In some situations, the ground truths may be inferred even though the model is complex.
    For example, \citet{poerner-etal-2018-evaluating} work on the subject-verb agreement task, predicting if the verb of a sentence is plural or singular based on the preceding texts. 
    Because the LSTM model they use achieves beyond 99\% accuracy, they infer that the model is always looking at the right word in the sentence (i.e., the corresponding subject) when making a prediction.
    As this subject can be accurately identified using a syntactic parser, they use the subject as an inferred ground truth which the explanation should assign the highest relevance score to. 
    
    \item \textbf{Unknown ground truths.} 
    Without any ground truths, some evaluation methods, instead, check whether the target model behaves in the same way as the explanation says.
    \citet{pljthesis} considers this as a looser definition of faithfulness.
    To evaluate relevance scores, for example, we can gradually remove or replace input words starting from the ones with high relevance scores to the ones with low relevance scores and then observe how the predicted probability changes \cite{arras-etal-2016-explaining,nguyen-2018-comparing,kim-2020}. 
    If the probability drops significantly when the words with high relevance scores are perturbed, we can consider the relevance scores faithful. 
    \textit{The area over the perturbation curve (AOPC) score}, which is a metric introduced by \citet{samek-2017} in the domain of computer vision, can be used to quantify faithfulness for this evaluation method.
    A very similar approach is called \textit{degradation tests} \citep{schulz-2020,chen-2021}. These additionally consider the least relevant perturbations and are based on the area between the perturbation curves.
    Instead of observing predicted probability, \citet{chen-ji-2020-learning} measure how many of the class predictions remain the same if only $k$ most relevant tokens (according to the relevance scores) is kept in the input. This results in a metric called \textit{post-hoc accuracy}.
\end{itemize}

In any case, because faithfulness concerns the accuracy of the explanations with respect to the underlying model, \citet{jacovi-2020} emphasize that we should not involve humans into the evaluation process.

\paragraph{Plausibility} \label{Plausibility}
In contrast to faithfulness, plausibility involves humans into the evaluation process.
Specifically, an explanation is plausible if it is convincing to humans towards the model prediction \cite{jacovi-goldberg-2021-aligning}.
Even an explanation for an incorrect prediction or an explanation that is not faithful to the underlying model can be plausible as long as human judgments say that it supports well the model prediction.
Therefore, one way to measure plausibility of an explanation is to compare it with human explanation(s). 
A machine explanation that has higher correlation with human explanation(s) is considered more plausible.
This idea was implemented by \citet{mohseni2018human}.
As another example, \citet{Lertvittayakumjorn-2019} select samples where a well-trained model has a high classification confidence. Then, for each of the selected samples, they apply an explanation method and present the most important features to humans (without showing the input text). If the humans correctly predict the output of the model, it implies that the explanation is plausible and well justifying the model output from human perspectives.

\section{Explainable MT evaluation} \label{sec:xmte}
In this section, we discuss main goals of explainable MT evaluation and define core properties that set this task apart from other tasks.
\subsection{Goals of explainable MT evaluation}
Different goals are sought to be fulfilled by explainability techniques \citep{lipton-2016,carvalho-2019,barredo-arrieta-2020} as discussed in the Introduction section.

\begin{table}[!htb]
\centering
    \begin{tabularx}{\textwidth}{lll}
     \toprule \textbf{Description} & \textbf{Use Cases} & \textbf{Subgoals} \\ 
     \midrule \makecell[l]{Diagnose and \\improve metrics} & \makecell[l]{- Build better MT systems} & \makecell[l]{- Informativeness \\ - Transferability} \\
     \midrule \makecell[l]{More expressive and \\ accessible metrics} & \makecell[l]{- Build researcher's trust\\ - Metric usage in new scenarios} & \makecell[l]{- Accessibility \\ - Transferability} \\
     \midrule \makecell[l]{Support Semi-Automatic\\ labeling} & \makecell[l]{- Faster annotation} & \makecell[l]{- Informativeness \\ - Interactivity} \\
     \midrule Checking for social biases & - \makecell[l]{Ensure fair MT models \\and metrics} & \makecell[l]{- Fair \& Ethical \\$\;\;$Decision Making\\ - Informativeness \\ - Trust}  \\
     \bottomrule
\end{tabularx}
\caption{Goals of explainable Machine Translation evaluation. The subgoals are described by \citet{lipton-2016,barredo-arrieta-2020}.}
\label{table:xMTEGoals}
\end{table}

\citet{lipton-2016} and \citet{barredo-arrieta-2020} define common goals in the development of explainable systems:  \textit{Informativeness}, \textit{Transferability}, \textit{Accessibility}, \textit{Interactivity}, \textit{Fair \& Ethical Decision Making} and \textit{Trust}. \textit{Informativeness} seeks to increase the amount of information conveyed by a model and about its solution process. 
\textit{Transferability} is the goal of applying a model to new domains. 
\textit{Accessibility} aims to enable better usage by non-experts. 
\textit{Interactivity} seeks to improve user-experience by using a model in an interactive setup. 
\textit{Fair \& Ethical Decision Making} is the goal of providing bias-free systems. 
Lastly, \textit{Trust} is an often subjective goal of achieving trust in a models behavior. 

Here, we consolidate four main goals for explainable MT evaluation and associate them with the common goals by \citet{lipton-2016} and \citet{barredo-arrieta-2020}, as displayed in Table \ref{table:xMTEGoals}.\footnote{As one main goal for explainable MT evaluation can associate with several common goals, we call the common goals subgoals in the table.} We follow with a short description of each of them below.  

\begin{itemize}
    \item \textbf{Diagnose and improve metrics}: By explaining why a metric predicted a certain score for a machine translation, where humans would assign a different score, the weaknesses of an existing metric might be understood. This might enable architectural changes or the selection of different datasets leading to metric improvement. Likewise, explaining the whole model can unveil if the metric follows desired general properties (see Section \ref{globalExpPaper}) or might otherwise be led astray by carefully crafted adversarial input. 
    This goal encapsulates sub-goals such as Informativeness and Transferability \citep[e.g.][]{lipton-2016, barredo-arrieta-2020}. 
    \citet{freitag-2020c} show that metrics can also include biases towards reference sentences, i.e., a preference towards REF, even though there are other correct translation options. They also state that references would often include \textit{Translationese}, i.e., source artifacts that originate when human translators do not find a natural translation. We count the quantification of these problems to diagnosis and metric improvement as well.
    \item \textbf{Make metrics more expressive and accessible}: A metric that assigns a score on a sentence level is difficult to understand for non-experts, e.g. non-native speakers. If a metric provides further information such as which words it considers incorrect, it may become more accessible. Similarly, for an expert it might save time and discussions to be presented with such explanations. This use case involves the goals of Informativeness and Accessibility \citep[e.g.][]{lipton-2016, barredo-arrieta-2020}. Further, we hypothesize that accessibility plays an important role in metric selection and might be one of the reasons for the broad usage of BLEU. In other words, researchers might have built trust in BLEU due to its long and wide usage. Hence, when explanations help build trust into black-box metrics, the paradigm shift to newer metrics might be propelled.
    \item \textbf{Support Semi-Automatic labeling}: The manual labeling process of data is expensive. Thereby, fine-grained annotations like word-level translation error detection \citep[e.g.][]{fomicheva-2021} are specifically difficult to obtain. Many data-labeling platforms support semi-automated labeling,\footnote{e.g. \url{https://datasaur.ai/}} i.e.\ pipelines, in which human annotators only need to check and correct automatic predictions. Likewise, obtaining automatic explanations could boost efficiency of the annotators in this use case.
    \item \textbf{Checking for social biases}: Learned metrics might be biased towards certain types of texts used during training. Biases could be detected by observing explanations where sensitive attributes (e.g., gender, political or racial aspects) are flagged to be important  \citep{lipton-2016}. For example, explanations could show that a metric considers male names more important for a translation than female names in the same scenario. Mitigating such biases involves goals such as Fair \& Ethical Decision Making, Informativeness and Trust \citep[e.g.][]{lipton-2016, barredo-arrieta-2020}. 
\end{itemize}

\subsection{Properties of explainable MT evaluation 
}
\label{sec:propertiesOfExplainableMTE} 

\begin{table}[!htb]
\centering
    \begin{tabularx}{\textwidth}{ll}
     \toprule \textbf{Property} & \textbf{ Short Description} \\ 
     \midrule Multiple Inputs & \makecell[l]{MT metrics take interdependent hypothesis-\\ and source-/reference- sentences as input} \\
     \midrule Relation to translation errors & \makecell[l]{Explanations of MT metrics are related to \\ translation error detection} \\
     \midrule Multilingual Inputs & \makecell[l]{Reference-free evaluation is one of few tasks\\ with multilingual inputs} \\
     \midrule Output Scores & \makecell[l]{Machine translation metrics return numeric outputs\\ (in contrast to classification models)} \\
     \midrule Varying Scales & \makecell[l]{The distribution of metric scores can\\ vary across metrics} \\ \midrule
     \makecell[l]{Label preservation \\ $=$ Meaning preservation} & \makecell[l]{ Adversarial attacks need to \\ preserve meaning} \\
     \bottomrule
\end{tabularx}
\caption{Properties affecting the explainability of MT evaluation metrics.}
\label{table:xMTEProperties}
\end{table}

Next, we consider, which factors of MT metrics distinguish them from other metrics and models in general and discuss the impact this has on their explainability. Table \ref{table:xMTEProperties} shows an overview of these properties. Now, we describe them in more detail:
\begin{itemize}
        \item \textbf{Multiple Inputs}: MT metrics take multiple interdependent inputs (hypothesis, source and/or reference(s)). Machine translations should fulfill the properties of fluency and adequacy, and metrics should evaluate these \citep[e.g.][]{yuan-2021}, i.e. the translation should be fluent - without grammatical errors and easy to read -, and adequate - without adding, losing or changing information. Each word in a correct translation depends on the source sentence and each word in the source sentence should be reflected in the translation (not necessarily one-to-one). As reference sentences are also translations of the source, the contents of references should also be reflected in the hypothesis sentence and vice-versa. As MT metrics take multiple inputs, there are different options of which parts of the input are explained when local (see Section \ref{subsec:scopeexp}) explainability techniques are applied: 
    \begin{itemize}
        \item \textbf{Explaining with respect to the Hypothesis}: Usually when Machine Translation is evaluated, the reference and/or source sentences are already given, for example, by a human annotator. Then, the metric can be used to generate a score that answers the question ``Is the hypothesis good with respect to the given source or reference?''. Explaining with respect to the hypothesis would additionally (try to) answer the questions ``Why did the metric assign a certain score for this hypothesis with respect to the given source or reference?'' and ``How do changes to the hypothesis affect the score?''.  
        \item \textbf{Explaining with respect to Source/Reference}: This would answer the additional questions of ``Why did the metric assign a certain score for this source or reference with respect to a given hypothesis?'' and ``How can we change the reference/source to affect the score in certain ways?''. 
        Usually the source and/or reference(s) are considered the ground truth of an evaluation. Therefore, it is not necessary to change them, as long as the integrity of the data is not questioned. They can however be interesting for model diagnosis to see whether the explained metric considers the expected parts of a source/reference when grading the hypothesis.
        \item \textbf{Explaining with respect to all inputs}: The question of whether all inputs can be explained together largely depends on the choice of the explainability method. For example, for Input Marginalization \citep{kim-2020}, the result will be the same as when the Hypothesis and Source/Reference are explained solely. This is 
        because this technique explains each word solely while keeping all other parts of the input fixed.
        However, for additive feature explanation methods such as SHAP \citep{lundberg-2017} and LIME \citep{ribeiro-2016}, the goal is to find a contribution of each feature to the final score. Therefore, when a SRC/REF and HYP are explained together by this technique, the scores assigned to each of their tokens will contribute to the final score together. This is not desirable as it 
        might
        lead to corresponding tokens between HYP and SRC/REF to be assigned with contradicting feature importance (one will be positive while the other is negative). 
        Additionally, doing this would invert the baseline of the attributions. As an example, assume we want to explain the reference-based metric score for a reference and a hypothesis. When we explain only the hypothesis, as described in the first bullet point, a baseline contribution could be determined by removing all words from the hypothesis (depending on the technique). Hence, the metric score is probably low, as it compares an empty sentence to a non-empty sentence. However, when the feature importance of reference and hypothesis is determined together, the baseline would be determined by comparing an empty hypothesis to an empty reference, leading to a high baseline score. 
    \end{itemize}
    \item \textbf{Relation to translation errors}: As adequacy is a key goal of MT \citep[e.g.][]{yuan-2021}, metrics should have lower scores when it is violated, i.e. if there are translation errors. Hence, it is intuitive that explanations of performant metrics should consider translation errors as important (or harmful to a metric's score).
    \item \textbf{Multilingual Input}: Reference-free metrics have multilingual inputs. While there are other tasks with multilingual inputs, such as cross-lingual summarization \citep{zhu-etal-2019-ncls} and cross-lingual semantic textual similarity \citep{cer-etal-2017-semeval}, 
    it is not common to many tasks and therefore should be considered when explaining them.
    \item \textbf{Output Scores}: Most MT evaluation metrics return numeric output scores; however, many explainability techniques in NLP have originally been used with classification. The same is true for the closely related field of adversarial attacks (see our Section \ref{sec:AdversarialAttacks}). 
    While some of them can be adapted to explain numeric scores easily (e.g., LIME \citep{ribeiro-2016}), others require the definition of intervals for which certain conditions should be met (e.g., Anchor \citep{ribeiro-2018}).
    \item \textbf{Varying Scales}: As many metrics are based on learned models, the outputs often are not fixed to a range between 0 and 1. Therefore, they cannot be directly interpreted as probabilities. This problem can be tackled by using an approach detailed in \citet{zhang-2020}. Here, the authors determine a lower boundary of the metric by averaging the scores of a large collection of dissimilar sentences. Then, they rescale all further computations of the metric with this baseline. However, while this leads to the score to be between 0 and 1 most of the times, it does not guarantee for it. 
    
    Another option is to normalize the scores to have a mean of 0 and a standard deviation of 1 (z-score) \citep{kaster-2021}. 
    For a given dataset this guarantees that scores are centred around 0 and spaced comparably across metrics. However, the normalized scores of new or perturbed sentences might follow a different distribution and fall out of this range. Additionally, the outputs of the metrics could be skewed differently. For instance, for two metrics that return outputs between 0 and 1, one could be more pessimistic and usually assign lower scores than the other. This makes comparison difficult as, for example, a score of 0.7 can have a different meaning for two metrics.
    \item \textbf{Label preservation $=$ Meaning preservation}: In general, adversarial attacks are label-preserving modifications of an input (designed to expose limitations of models). In MT, the concept of label-preservation coincides with the notion of meaning-preservation (via the concept of `adequacy' that MT metrics typically measure). This requires stronger adversarial attack methods; see our Section \ref{sec:new}.   
\end{itemize}


\section{Datasets}\label{sec:datasets} 

Explainability datasets usually include human explanations for a certain task
(e.g., text classification). The plausibility of machine generated explanations can then be evaluated by comparing them against the ground truth.\footnote{In this Section we focus on the plausibility aspect of explanations, see Section \ref{EvalOfExplanations} for the discussion of faithfulness.} Three types of ground truth explanations are typically considered: \textit{highlights} (also known as rationales), \textit{free-text} explanations and \textit{structured explanations} \cite{wiegreffe-2021}. 
Based on the terminology we introduced in Section \ref{sec:xai}, 
highlights are \emph{feature importance explanations}, where 
the most important tokens are selected as an explanation. 
Free-text explanations and structured explanations are
both \emph{textual explanations}, but structured explanations have constraints
on the form of the text. 

With the exception of the recent work by \citet{fomicheva-2021}, there are no datasets devised explicitly for studying explainability of MT evaluation metrics. However, depending on the type and definition of explanations some of the existing datasets for MT evaluation can be leveraged for this purpose.

As mentioned in Section \ref{sec:propertiesOfExplainableMTE}, one of the interesting properties of the MT evaluation task is that erroneous words in the MT output serve as explanations for sentence-level or system-level quality scores. Table \ref{tab:mt_explanations_example} provides an example of the three types of explanations mentioned above for sentence-level MT evaluation.

\begin{table}[ht]
    \footnotesize
    \centering
    \begin{tabular}{p{0.35\linewidth} | p{0.6\linewidth}}
      \toprule
      Data & Explanation \\ \midrule
      \textit{(Source)} Pronksiajal voeti kasutusele pronksist tooriistad, ent kaepidemed valmistati ikka puidust \textit{(MT)} Bronking tools were introduced during, but handholds were still made up of wood.\newline Score: 0.3 & \textbf{Highlights}:       \textit{(Source:)} \hl{Pronksiajal} voeti kasutusele \hl{pronksist} tooriistad, ent \hl{kaepidemed} valmistati ikka puidust \textit{(MT)} \hl{Bronking} tools were introduced during the \hl{long term}, but \hl{handholds} were still made up of wood\newline \textbf{Free-text}: MT quality is low because it contains two lexical errors and one omission error.\newline \textbf{Structured}: 2 lexical errors, 1 omission error \\ \bottomrule
    \end{tabular}
    \caption{Example of highlight-based, free-text and structured explanations for MT evaluation of a sentence translated from Estonian into English.}
    \label{tab:mt_explanations_example}
\end{table}

Defining explanations for MT evaluation as translation errors allows to leverage a wide variety of existing MT evaluation datasets to study explainability. 

\paragraph{Post-editing based datasets}
A popular method for obtaining silver labels for MT evaluation is post-editing. Post-editing based datasets contain sources sentences, machine translations and the corresponding human post-edits (PEs) \cite{fomicheva-2020}. Sentence-level quality scores are computed as the minimum distance between the MT and the PE (the so called HTER score \cite{snover2006study}), whereas word-level labels can be derived from the same data by computing the minimum distance alignment between the MT and its post-edited version. Thus, the misaligned words in the MT naturally provide a ground truth explanation for the sentence-level score.  
An important disadvantage of these datasets, however, is the noise introduced by the automatically computing the MT-PE alignment. 

\paragraph{Error annotation}\label{MQM}
Another type of dataset where both word and sentence-level scores are available are the datasets based on manual error annotation. MT error annotation protocols such as MQM (Multidimensional Quality Metrics) \citep{lommel-2014, lommel-2015} frame MT evaluation as an error annotation task. Each word in the MT output is assigned a quality label based on a fine-grained error typology. The sentence-level score is then derived as a weighted average of word-level errors. Thus, in this case the error labels can be used as explanations for a metric that outputs sentence-level MQM scores. Note that this type of data would also allow for explanations that involve the type of error, as illustrated in the free-text and structured explanation in Table \ref{tab:mt_explanations_example}. The disadvantage of this annotation scheme is low inter-annotator agreement and high annotation costs.

\paragraph{X-QE dataset}
Few datasets employ human annotated rationales as explanations for machine translation evaluation. The Eval4NLP 2021 shared task \citep{fomicheva-2021,eval4nlp-2021-evaluation} provided the first dataset that jointly annotated sentence-level scores with word-level error highlights (seen as explanations for the sentence-level scores) for the MT setting. Another related dataset consists of annotations collected in the domain of crosslingual divergence \citep{briakou-2020}.

Overall, existing datasets can be used to evaluate the plausibility aspect of explanations for MT evaluation systems by leveraging the relation between word-level and sentence-level quality. However, the error-based definition discussed in this Section has two important limitations. On the one hand, it is not clear what the highlight-based explanations should look like for high-quality MT. On the other hand, it is not clear what the explanations should look like for an MT evaluation system that operates at word level (i.e. predicts translation errors).

\section{Taxonomy of Existing Explainable Evaluation Metrics} \label{sec:taxonomy}
\begin{table*}[!htb]
\small 
\centering
\begin{tabular}{p{5.5cm}ccccccc}
     \toprule
     \textbf{Work} & \textbf{Type} & \textbf{L/G} & \textbf{Method} & \textbf{A/S} & \textbf{Goals}\\ \midrule
     Eval4NLP 2021: \citep{fomicheva-2021} &  &  & Various &  & \\\hdashline\noalign{\vskip 0.5ex}
     \;\citet{rubino-2021} & FI & L & Expl. by Design & S & A/L\\\hdashline\noalign{\vskip 0.5ex}
     \;\citet{treviso-2021} & FI & L & Various & A/S & A/L\\\midrule
     SemEval 15/16: \citep{agirre-2015,agirre-2016} & & & Various & &\\\hdashline\noalign{\vskip 0.5ex}
     \;\citet{magnolini-2016} & CAl & L & Neural Netw. & S & A/E
     \\\midrule
     \citet{yuan-2021} & CA & L & Generation Prob. & S & E \\\midrule
     Adversarial Attacks (Section \ref{sec:new}) & EbE & L & Perturbations & A & D/E \\\midrule
     \citet{kaster-2021} & EbS/CA & G & Linear Regr. & A & D/E \\\midrule
     \citet{sai-2021} & CA & G & Perturbations & A & B/D/E \\
     \bottomrule
\end{tabular}
\caption{Explainability for MT metrics. We distinguish the explanation types Concept Attribution (CA), Chunk Alignment (CAl), Feature Importance (FI), Explanation by Example (EbE) and Explanation by Simplification (EbS). L/G considers Local (L) vs Global (G) and A/S considers (A)gnostic vs (S)pecific. The column ``Goals'' specifies which of the goals of Section \ref{sec:xmte} can be addressed with the respective techniques. Thereby, we consider (B)ias detection, metric (D)iagnosis, metric (E)xpressiveness and automated (L)abeling.}
\label{taxonomy}
\end{table*}


In this section, we categorize and summarize recent approaches 
related to the concept of explainable MT evaluation. Based on the dimensions we introduced in Section \ref{sec:xai}, we 
describe
the techniques themselves 
and discuss how they relate to the properties of explainable MT evaluation introduced in Section \ref{sec:xmte}. In Table \ref{taxonomy}, 
we provide an overview table of our taxonomy. 



\subsection{Local Techniques}
\todo{SE: agreement in our discussion group on Friday: just briefly described Eval4NLP approaches, more in a summarizing form}
As described in the previous section, local explainability techniques provide additional information that helps 
understand model behavior with specific input/output pairs. 

\paragraph{Word-Level Feature Importance Explanations}

When humans evaluate translations, they often focus on the errors that they can identify on a word- or phrase-level \citep{freitag-2021}. \citet{fomicheva-2021b} and the 2021 Eval4NLP shared task \citep{fomicheva-2021}
build on this idea and evaluate how well feature importance is correlated with human word-level error annotations. 
They use this correlation as an indicator of plausibility (see Section \ref{Plausibility}). This follows the intuition that showing word-level errors as an explanation is plausible to humans, who look for the same kind of clues. 
In other words, word-level error extraction from sentence-level metrics could be used as additional benchmark for explainability methods. 
\citet{fomicheva-2021b} explore this approach with the supervised reference-free metric TransQuest \citep{ranasinghe-2020}. 
They manually label correct words with 0 and incorrect words with 1 in the MT hypothesis. As feature importance scores are continuous rather than binary, they use the metrics \textit{area under the receiver operator characteristic curve} (AUC), \textit{average precision} (AP), \textit{recall at top-k} (R@K) and \textit{accuracy at 1} (A@1) to compare their manual annotation to automatic feature importance scores. 
\citet{fomicheva-2021b} choose four 
well-known explainability techniques in order to extract the feature importance scores, (i) \textit{LIME} \citep{ribeiro-2016}, (ii) information bottleneck \citep{schulz-2020}, (iii) integrated gradients \citep{sundararajan-2017} and (iv) attention \citep[e.g.][]{wiegreffe-2019, serrano-2019}. 
\todo{SE: I commented out the below - maybe we explain LIME, etc. in Section 3?}
Additionally, \citet{fomicheva-2021b} compare a metric that was trained with supervision to compute word-level errors in a classification setting \citep{ranasinghe-2021} to a glass-box method that uses the word-level translation probabilities of each word of the known MT model as feature importance \citep{fomicheva-2020b}.\todo{SE: I don't understand what two things are compared: what is a glass-box method? Is it the above LIME, etc. stuff?}
The following key findings are reported:
\begin{itemize}
    \item LIME performs worse than the other methods, which is attributed to the fact that LIME works with perturbations of the input rather than at a hidden state. Perturbations on input are not suitable method when explaining MT evaluation since removing an erroneous word does not make the sentence correct. 
    \item Feature importance for word-level explanation performs competitively to the glass-box method, and integrated gradients even approaches
    the performance of the supervised metric.
\end{itemize}
\todo{SE: it's not entirely clear to me what is going on, maybe make a graphical illustration? --> \\CL: I'll search the graphic from Marina's paper for tomorrows presentation, such that we can have a look at that before I create one to put in this place}

\label{eval4nlp}
The 2021 Eval4NLP \citep{fomicheva-2021} shared task explores 
a very similar 
evaluation approach of plausibility 
as \citet{fomicheva-2021b}. 
For training and development phases, they provide extracts from MLQE-PE. However, as test set they provide a novel dataset that contains directly annotated word-level labels (see Section \ref{sec:datasets}). As baseline systems, the organizers provide a random baseline, as well as TransQuest explained with the model-agnostic LIME and XMoverScore \citep{zhao-2020} explained with the model-agnostic SHAP \citep{lundberg-2017}. Seven systems were submitted to the shared task, three of which leveraged word-level supervision: one system with synthetic data \citep{rubino-2021} and two with manually annotated data \citep{treviso-2021,polak-2021}. In the following, we present a short summary of the two best performing submissions:\todo{SE: better we make a table with all systems and briefly categorize them. Then we just highlight a 3-4 interesting approaches}



\begin{itemize}
\item
\textit{Error identification with metric embeddings and attention.} The approach by \citet{rubino-2021} jointly fine-tunes an XLMR model with a so-called metric embedding matrix. For each pair of input sentences, multiple standard metrics (e.g.\ BLEU and CHRF) are computed\todo{SE: how can they use BLEU for source and hypothesis?} and the resulting scores are multiplied with the matrix. This leads to a metric embedding, a learned vector representation of the metric results. They then leverage the metric embedding in an attention mechanism with the hidden states of the XLMR model in order to learn which parts of an input which metrics focus on. They then leverage the attention weights in the computation of sentence- and word-level scores.\todo{SE: they won the shared task, but used synthetic data} This approach is a local explainability technique that is explainable by design. 



\item \textit{Scaled attention weights.} \citet{treviso-2021} fine-tune 2 XLMR models \citep{conneau-2020} and a RemBERT \citep{chung-2021} for sentence-level quality estimation. They then explore various explainability techniques to extract meaningful word-level scores. Specifically, they explored 3 attention mechanisms, (a) using the row-wise average across all attention heads, (b) only averaging promising rows, (c) scaled by the norm of value vectors \citep{kobayashi-2020}. They also explore gradient-based methods (i) using the gradient of hidden states multiplied with the hidden states, (ii) using the gradient of hidden states multiplied with the attention weights, (iii) using Integrated Gradients \citep{sundararajan-2017}. 
Finally, they employ a method that learns binary masks of input features by using a second model \citep{bastings-2019}. They achieve their best results for an unsupervised setting with the normalized attention.\todo{SE: best explainability approach}
\end{itemize}




\todo{SE: mention evaluation from Section 7}
\limit{
First, a potential issue with the evaluation approach of Eval4NLP is that it does not consider the property of faithfulness (see Section \ref{Faithfulness}). Hence, there is no guarantee that the extracted word-level scores actually reflect the sentence-level score, i.e., explain the sentence-level metric scores. 
Second, certain translation errors cannot be easily captured by highlighting specific words. For example, the annotation scheme cannot handle cases where the MT fails to explicitly express a grammatical feature that is omitted in the source language but is required to be explicit in the target language (e.g.\ the use of articles when translating from Russian into German). Third, different translation errors affect the sentence score to a varying extent, which cannot be properly captured with binary labels. 
Fourth, the approach in Eval4NLP does not provide correspondence between highlighted error words in the source and target language.  
Finally, ranking the participating systems according to their global independent statistics might be unreliable, as we discuss in Section \ref{sec:bt}.}



\paragraph{Chunk Alignment} 
For the field of \textbf{semantic textual similarity} (STS), which is very 
closely related to evaluation metrics, more fine-grained forms of local explainability have been explored. 
The second tasks of SemEval-2015 \citep{agirre-2015} and -2016 \citep{agirre-2016} 
ask participants to perform a labeled chunk alignment between two sentences as explanation for STS. 
In the respective datasets, they annotate how phrases relate between sentences and assign scores to the relation strength based on a 0 to 5 scale. Also, they assign labels such as ``similar'' to define the type of the relation. 
As example, the authors consider the sentences 
\begin{align*}
    s_1 &= ``\text{12 killed in bus accident in Pakistan}''\\ 
    s_2 &= ``\text{10 killed in road accident in NW Pakistan}''
\end{align*} 

They show sample alignments for ``[12] $\Leftrightarrow$ [10]'' to be ``similar'' with a relation strength of 4 and ``[in bus accident] $\Leftrightarrow$ [in road accident]'' to be a ``specification'' with relation strength of 4. The annotation process first assigns chunks, then relation strength and then the label(s).
To measure the system quality, they use different F1 measures taking into account the scores and labels. Their baseline system uses rules to assign the label. The best performing system at that time used a multi-layer perceptron and multi-task learning \citep{magnolini-2016}. There are recent works that improve on the task, especially the phrase-level alignment. For example \citet{lan-2021, arase-2020} provide new datasets and approaches for this topic. 

\limit{The approach requires more fine-grained annotation, which would result in lower agreement levels among annotators or less reliable automatic annotation. The relation of the annotation to translation errors, as a key factor in explainable MT metrics, is also not explicit in the scheme.}

\paragraph{Generation Direction}
The recently proposed metric BARTScore by \citet{yuan-2021} treats NLG evaluation as a generation task. They use BART \citep{lewis-2020} for text generation to predict the probability of a sentence to be generated given another sentence. The resulting probability is used as metric score. 
\citet{yuan-2021} evaluate different generation directions: Faithfulness: SRC$\rightarrow$HYP, Precision: REF$\rightarrow$HYP, Recall: HYP$\rightarrow$REF, F-Score: HYP$\leftrightarrow$REF (the arithmetic average of precision and recall). For example, for SRC$\rightarrow$HYP the metric score is the probability that BART generates the HYP given the SRC. 
They state that these directions would encapsulate different evaluation perspectives. E.g.\ HYP$\rightarrow$REF would encapsulate \textit{semantic coverage}, i.e.\ how well the meaning of the reference is captured by the hypothesis. As such, providing the results of different generation directions can be treated as explainability approach that provides additional information on a sample instead of a single metric score. 

Based on the evaluation directions presented by \citep{yuan-2021}, the recent ExplainaBoard webtool \citep{liu-2021} provides a possibility to compare and benchmark evaluation metrics. This dimension is provided under the title ``Meta Evaluation for Automated Metrics''. 

\limit{The provided scores of different generation directions may not readily be meaningful to different users.}

\subsection{Global Techniques} \label{sec:global_techniques}
As discussed in the previous sections, most explainability techniques for neural networks consider local explanations. While these can provide an insight into a model's decision process for specific samples, often it also is desirable to characterize how the model will behave in general. In this section, we consider recent approaches of global explainability.

\paragraph{Disentanglement along Linguistic Factors}
\label{globalExpPaper}
\citet{kaster-2021} propose a global explainability technique that decomposes the score of sentence-level 
BERT-based metric into different linguistic factors. In particular, they explore the degree to which metrics consider each of the properties \textit{syntax}, \textit{semantics}, \textit{morphology} and \textit{lexical overlap}.  
As a first step, they 
define several explanatory variables: 
\begin{itemize}
    \item \textbf{Lexical Overlap Score (LEX):} The lexical overlap score between a hypothesis and a reference/source is determined using BLEU \citep{papineni-2002} with unigrams, 
    thus ignoring word order. 
    For reference-free metrics, they generate pseudo-references by translating the source sentences with Google Translate before applying BLEU.
    \item \textbf{Morphological Score (MOR):} The morphological score computes the cosine similarity of morphologically enriched sentence embeddings. These are based on word embeddings 
    finetuned on morphological lexicons following \citet{faruqui-2015}.
    \item \textbf{Semantic Score (SEM):} The semantic score leverages human annotated sentence level scores (adequacy or semantic similarity) provided for different datasets. 
    \item \textbf{Syntactic Score (SYN):} The syntactic score measures the syntactic similarity between hypothesis and source/reference with the tree edit distance (TED) \citep{bille-2005} of parse trees generated by the Stanford dependency parser \citep{chen-2014}.
\end{itemize}
\citet{kaster-2021} apply z-score normalization to make the scores of these explanatory variables comparable. Table \ref{table:example} shows an example they use to demonstrate the different scores for two sentence pairs. 

\begin{table*}[!htb]
\setlength{\tabcolsep}{3.5pt}
\small 
\centering
\begin{tabular}{p{4.5cm}|p{4.5cm}|cccc}
     \toprule
     \textbf{Hypothesis} ($y$) & \textbf{Reference}/\textbf{Source} ($x$) & \textbf{SEM} & \textbf{SYN} & \textbf{LEX} & \textbf{MOR}\\ \midrule
     It is a boy , likes to sport , but it cannot do it because of their very. & He is a boy, he likes sports but he can't take part because of his knee. &  -1.57  & 0.98 & -0.59 & -0.87 \\ \midrule
     
     Zwei Besatzungsmitglieder galten als vermisst.	 & Two crew members were regarded as missing.  &	0.83 & 0.99  & 0.46 & -2.40\\
     \bottomrule
\end{tabular}
\caption{Example setups with normalized semantic, syntactic, lexical overlap and morphological scores \citep{kaster-2021}.}
\label{table:example}
\end{table*}
%
They then estimate the following linear regression:
$$m(x,y)=\alpha\cdot \textit{sem}(x,y)+\beta\cdot \textit{syn}(x,y)+\gamma\cdot \textit{lex}(x,y)+\delta\cdot \textit{mor}(x,y)+\eta$$
where $x$ is a reference or source sentence, $y$ is a hypothesis sentence and $sem$, $syn$, $lex$ and $mor$ are the scores of the respective linguistic properties. $m$ is the metric that is explained, $\eta$ is an error term and $\alpha$, $\beta$, $\gamma$ and $\delta$ are learned weights that indicate the linear influence each property has on the metric score. The degree to which the learned regressors approximate the metric function is determined with the determination coefficient $R^2$.
%
%

They conduct experiments with data from WMT15-WMT17 \citep{stanojevic-2015,bojar-2016,bojar-2017} for the domain of machine translation and STSB \citep{cer-2017} for the domain of semantic textual similarity. The following key findings are reported:
\begin{itemize}
    \item The $R^2$ value is generally higher for reference-based metrics than for reference-free metrics, implying that the learned linear regression can better explain reference-based metrics. \citet{kaster-2021} hypothesize that this is due to missing explanatory variables (regressors) or non-linear relationships. 
    They introduce a fifth property, \textit{cross-lingual bias}, whose ``ground truth'' is given by the scores a reference-free metric returns in a reference-based setup. The reference is computed with Google Translate. Including this factor with an additional regressor improves the $R^2$ score however only in some settings.  
    \item Each metric captures semantic similarity and lexical overlap to some degree. Syntactic and morphological similarity are either captured to a smaller extent or are even negatively correlated with the metric score. Especially, MoverScore and BERTScore have a comparatively high coefficient for the lexical score.
    \item Based on the finding that the metrics favor lexical overlap, \citet{kaster-2021} explore their robustness towards particular adversarial examples that preserve lexical overlap but change meaning. They show that non-paraphrases that have a high lexical overlap but do not preserve semantics tend to achieve better scores 
    than paraphrases with low lexical overlap. This exposes an important weakness of the metrics.  
\end{itemize}

In terms of the dimensions introduced in Section \ref{sec:xai}, we can further categorize this explainability technique as ``explanation by simplification'', as a simple linear model is learned that explains the complex metric score as a linear combination of scores that can be more easily understood. We also point out that it can be used as a local explainability techniques, as it also explains individual instances. 

\limit{
In terms of goodness-of-fit, the approach could not explain 
reference-free metrics well, so plausibly requires alternate explanatory variables.  
The search for such regressors may be inspired by other text generation tasks (such as summarization) where not a global metric score is reported but several scores (such as coherence, fluency, etc.). 
These could then decompose a global MT score; 
cf.\ the discussion in \citet{sai-2021} where they argue against one global score for evaluating NLG. 
Further, 
SEM, which is determined by humans, and MOR, which is based on word embeddings, could be considered black box variables themselves and future work might replace them by more transparent factors. 
One might also explore the collinearity of the different regressors 
and define regressors more plausbibly (especially MOR, which concurrently captures both semantic and morphological aspects). 
}

\paragraph{Peturbation Checklists}
\label{sec:pertChecklist}
\citet{sai-2021} start out with the premise that evaluation in text generation is a multi-faceted problem that cannot be captured, in general, by one overall score. Instead, they 
propose perturbation checklists that evaluate how susceptible metrics are towards predefined types of permutations as evaluation criterion. 
This allows developers to check whether all invariances that are required for a specific task are fulfilled. 
Since the goal is then to evaluate a whole metric using perturbation templates, we classify it as a global technique. We point out, however, that their approach could also explain individual instances, just like the approach of \citet{kaster-2021}. 

To evaluate each metric with respect to different properties, they follow \citet{ribeiro-2020} and check how a metric behaves when property specific changes are applied to the input. In particular, they compare the change in the metric's score with the change in score a human would assign after the perturbation:  
\begin{align}\label{eq:sai}
    s(m) = \bigl( h(\hat{p}) - h(p) \bigr) - \bigl( f_m(\hat{p}) - f_m(p) \bigr)
\end{align}
Here $s$ denotes the score for a perturbation template $t$ that applies criterion (property) $c$. $m$ denotes the metric that is explained, $h$ is the human score and $f_m$ 
the score achieved by $m$. Finally, $p$ is the original model output and $\hat{p}$ the 
output after permutation. They annotate $h$ using 15 annotators that determine, on a scale from 0 to 10, how much a score should change for a perturbation. In total, they provide 34 perturbation templates. 
For MT, these templates encompass dropping or adding of context as well as negations. For other tasks, they have other templates, such as ones involving fluency or correctness (e.g., with respect to gender). 
They report the following key findings:
\begin{itemize}
    \item 
    BERTScore, BLEURT and Moverscore are shown to have problems to correctly predict the score with antonyms and negations. This may not be surprising given that BERT representations may be similar for antonyms. 
    \item The perturbation checklists allow to pick metrics that are strong with respect to specific properties. E.g., Moverscore would capture fluency better than BERTScore due to its ability to cope with the jumbling of words. 
    \item  Overall, across metrics and NLG tasks, they show that existing evaluation metrics are not robust against many simple perturbations and disagree with scores assigned by humans. 
    This allows for more fine-grained assessment of automatic evaluation metrics which exposes their limitations and 
    supports the design of better metrics. 
\end{itemize}

Following the definitions in Section 
\ref{sec:xai}, 
the perturbation checklists are model-agnostic and 
advance 
all three goals: they can detect biases and model shortcomings by applying respective perturbation templates;  further, they might improve accessibility in terms of improving the understanding what a metric does. However, as it is a global technique, it does not support users in understanding single decisions. 

\limit{A drawback of the approach is the need to craft specific templates and the associated human annotation effort. Automatizing this process would be highly desirable, cf.\ our novel approaches in Section \ref{sec:new}. Another problem in Eq.~\eqref{eq:sai} above is that metrics may have different scales, i.e., a metric may have a very narrow range and thus small deviations in general. This may yield misleading conclusions when compared to the human scale of annotations (the paper addresses this by normalizing all scores). 
Varying scales are 
also one reason to choose preferences over Likert scales in annotation tasks  \citep{kiritchenko-mohammad-2017-best}.}

\section{New explainability approaches for MT evaluation} \label{sec:new}
\todo{SE: make numbers in figure bigger and fix varying notation conventions \\CL maybe I can do it tomorrow}



In this section, we present new results and techniques and discuss their implications on explainable MT evaluation. 
First, in Section \ref{sec:AdversarialAttacks}, we analyze whether adversarial attacks can automatically spot weak spots in evaluation metrics, thereby contributing to their explainability especially from a system developer perspective.  
Then, in Section \ref{sec:hard}, we discuss novel simple explainable metrics for the Eval4NLP shared task and show that they can achieve strong results, which  
confirm the issue that the Shared Task's setup does not test the faithfulness of explanations. 
Finally, in Section \ref{sec:bt}, we analyze system comparisons for the Eval4NLP shared task. This is an important issue, as it contributes to the evaluation of explainability approaches for MT metrics, a currently neglected research area. 

\subsection{Adversarial Attacks}\label{sec:AdversarialAttacks}

\citet{szegedy-2014, goodfellow-2015} show that neural networks are susceptible to \emph{adversarial attacks} --  
minimal label-preserving perturbations of the input 
of deep learning models that may change their output.  
In object recognition for example, inputs can be augmented with noise that is imperceivable for humans but leads to different prediction outcomes \citep{szegedy-2014, goodfellow-2015}. 
The NLP community has 
differentiated between sentence-, word- and character-level attacks \citep{zeng-2020}. 
Sentence- and word-level attacks often have the goal to produce examples that are similar in terms of meaning \citep{alzantot-etal-2018-generating,li-2020}, while character-level attacks 
mimick various forms of typographical errors (including visual and phonetic modifications) \citep{ebrahimi-2018,pruthi-etal-2019-combating,eger-2019,eger-benz-2020-hero,keller-etal-2021-bert}.  

Interestingly, there is a deep connection between explainability and adversarial attacks, which is underrecognized in the literature. For example, \citet{linardatos-2021} list adversarial attacks as tools for sensitivity analysis, one branch of methods they identify to explain black box models. 
Gradient-based explanation techniques identify salient words, while adversarial attacks 
target such vulnerable words \cite[][e.g.]{li-2020, jin-2020}. A controversly discussed topic is the close relation of adversarial examples and counterfactuals \citep{freiesleben-2021}. 
Also, presenting 
 adversarial examples for a particular input could be interpreted as a local explanation-by-example technique. 
Finally, knowing a model's failure cases -- as adversarial attacks reveal -- helps us better understand the model. 
As such, we interpret our below adversarial attack experiments on evaluation metrics as contributing to their explainability. \citet{sai-2021} and \citet{kaster-2021}   follow similar approaches in that they perform input perturbations from a manually selected range of options (e.g. negation, replacing named entities, ...). Our approach differs from the last two by instead leveraging adversarial methods such as BERT-Attack \citep{li-2020} 
to \emph{automatically} find failure cases. This could evaluate robustness and help understand model performance 
at a larger scale. 


\begin{figure}[ht]
    \centering
    \includegraphics[width=0.8\textwidth]{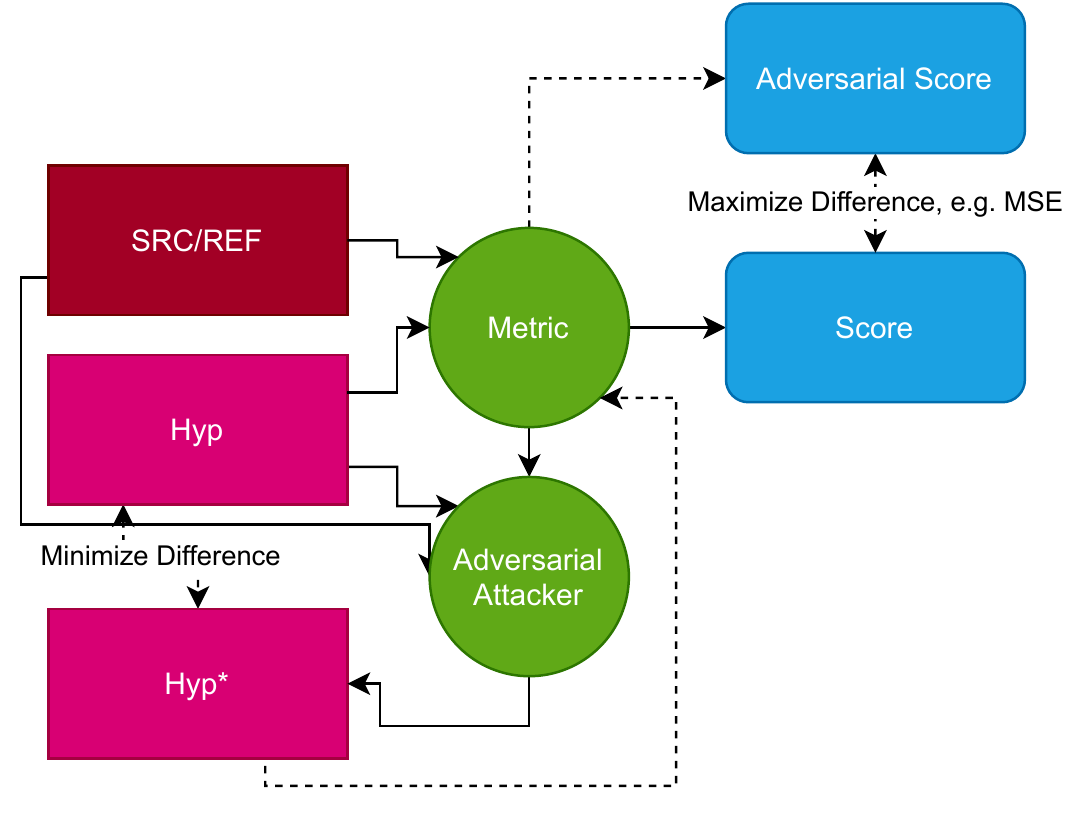}
    \caption{Schematic illustration of Adversarial Attacks on Machine Translation Evaluation Metrics.}\label{advDesign}
\end{figure}

A generic setup of adversarial attacks 
on evaluation metrics is shown in 
Figure \ref{advDesign}. 
Here, an adversarial attacker takes the original input of a metric and perturbs it to generate an adversarial hypothesis HYP$^*$. 
The metric then computes another score for the adversarial example. The attack is successful if this adversarial score is as much different from the original score as possible and the perturbed hypothesis is as much similar to the original hypothesis as possible. 
We note that the challenge lies in finding hypotheses HYP$^*$ that minimally deviate from the original hypothesis while maximizing the score differences. 

Most word- and sentence- level adversarial attacks consider a classification setting. To apply them to continuous metrics, we discretize the metric score into classes following a three step process. First, we select a dataset and calculate a metric score for each sample and metric. Second, we determine $k$ quantiles for each metric's scores, where $k$ is the number of classes we want to discretize to. Third, for each class of interest, we filter the metric scores for samples that lie in the same class for all metrics. Class probabilities are based on the procentual distance to the center of the class intervals the metric score lies in between. 
We apply the following adversarial attackers to the discretized scores:
\begin{itemize}
    \item \textbf{BERT-Attack} \citep{li-2020} leverages BERT \citep{devlin-2019} for word replacement in a word-level adversarial attack. First, it ranks tokens based on their importance for the original prediction (here the metric score) using feature importance explanations. Then the tokens are iterated in order of this ranking. For each token, $k$ replacement candidates are checked and if they are not successful in keeping the original prediction and a similarity constraint, the next token in the importance ranking is assessed. If the token is a word, it is replaced with one of the top-$k$ replacement candidates obtained by feeding the sentence to BERT (without usage of a mask). If the token is a sub-word, the other sub-word tokens that belong to the word are identified, all possible sub-word combinations are listed and their probabilities to occur as a word replacement are determined with BERT. Each replacement is evaluated with respect to the goal of the attack. Once the original prediction changes to another quantile, the attack is successful. When many sub-words occur in one word, the number of combinations can grow prohibitively large. In these cases, we cap computation at 500k combinations. 
    \item \textbf{Textfooler} \citep{jin-2020} also ranks the words by their feature importance. The ranked words are iterated and, for each word, $k$ replacement candidates are tested. The candidates are determined as the top-$k$ words that have the same POS-tag and whose static word embeddings have a high cosine similarity to the original word. 
    Further, only candidates that keep the universal sentence encoder (USE) \citep{cer-2018} similarity 
    to the original sentence higher than a threshold are considered. If the predicted class does not change after checking all candidates for a word, the candidate that minimized the predicted class probability is kept and the next important word is perturbed. 
    \item \textbf{Textfooler-Adjusted} \citep{morris-2020-reeval} adds further constraints to Textfooler. It sets higher thresholds on minimal word- and sentence-level similarity (based on static word embeddings and USE). Additionally, it employs 
    an open-source grammar checker 
    \citep{naber-2003} to impose a constraint on the number of grammatical errors introduced by an attack.
\end{itemize}

We use TextAttack \citep{morris-2020-textattack} as framework for adversarial attacks. 
The described attacks are originally \emph{untargeted}, i.e.\ they follow the goal of changing the resulting class no matter the direction. 
However, we use a targeted setup (with a desired target class specified) for BERT-Attack, Textooler and Textfooler-Adjusted, by leveraging the respective class of Textattack. 

\paragraph{Automatic Evaluation}
We apply attacks to a subset of the de-en (German-English) dataset of WMT19. 

We divide the scores into 3 classes and filter for all samples that each considered metric places in the same class. Improving translations is a difficult task compared to making them worse, hence we investigate the change from the highest class to the lowest class. 
There are 440 instances which fall in the highest class for all metrics.\footnote{We consider a mix of hard and soft reference-based as well as reference-free metrics.} 

\begin{figure}
  \centering
    \includegraphics[width=\textwidth]{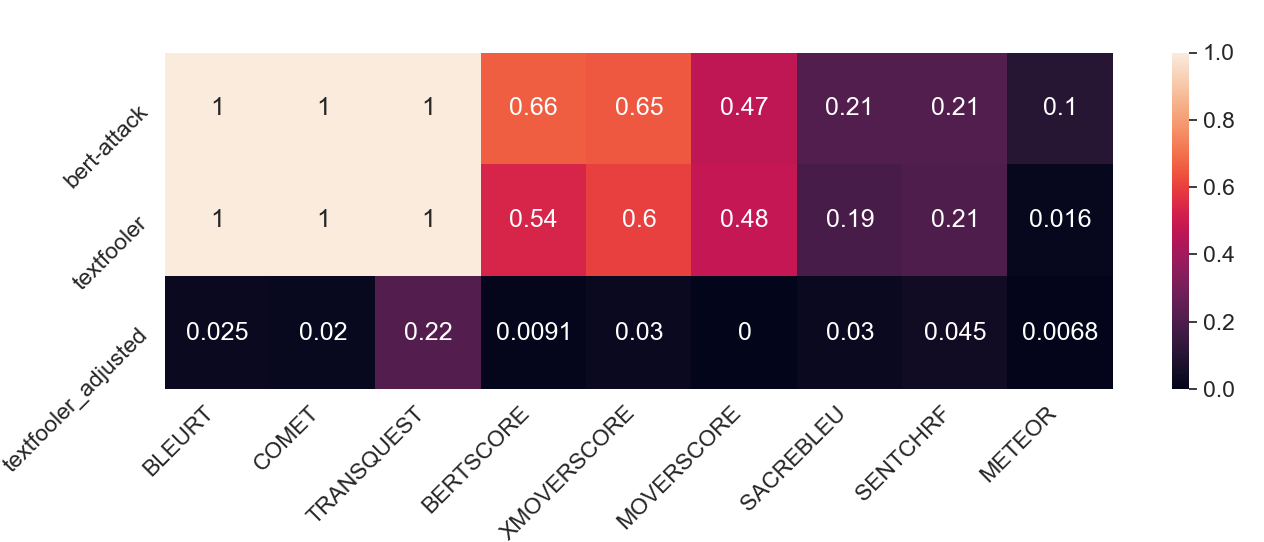}
    \caption{Success rate of adversarial attacks on 440 de-en samples from WMT19.}
    \label{fig:att-de-en-wmt19-success}
 \end{figure}

Figure \ref{fig:att-de-en-wmt19-success} 
shows the success rate of the attacks per metric, i.e., the percentage of
the time that an attack can find a successful adversarial example by perturbing a given input. A surprising finding is that BERT-Attack and Textfooler apparently perform very well against supervised metrics, have a smaller success rate on metrics 
based on word-level similarities such as XMoverscore and BERTScore and are least successful for hard metrics. 
This result is to some degree counter-intuitive, as hard metrics could apparently be fooled by simple lexical variations. 
The pattern is different for TextFooler-Adjusted, which has low success rates throughout, except for Transquest, where it has 22\% success rate (followed by hard metrics). 

\begin{figure}
  \centering
    \includegraphics[
    scale=0.385]{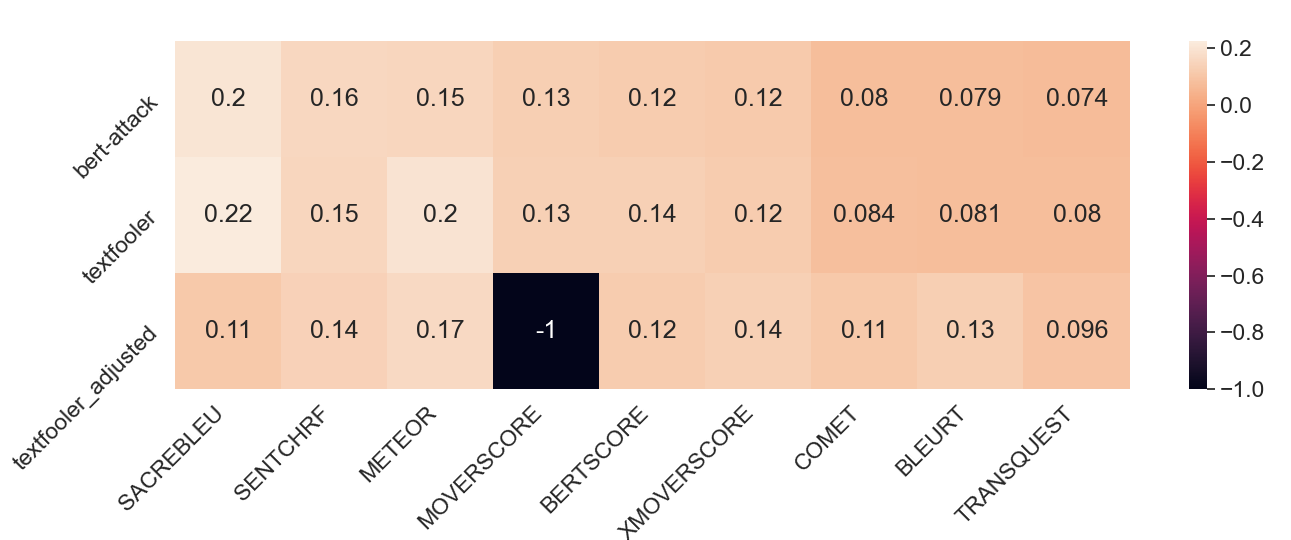}
    \includegraphics[
    scale=0.4]{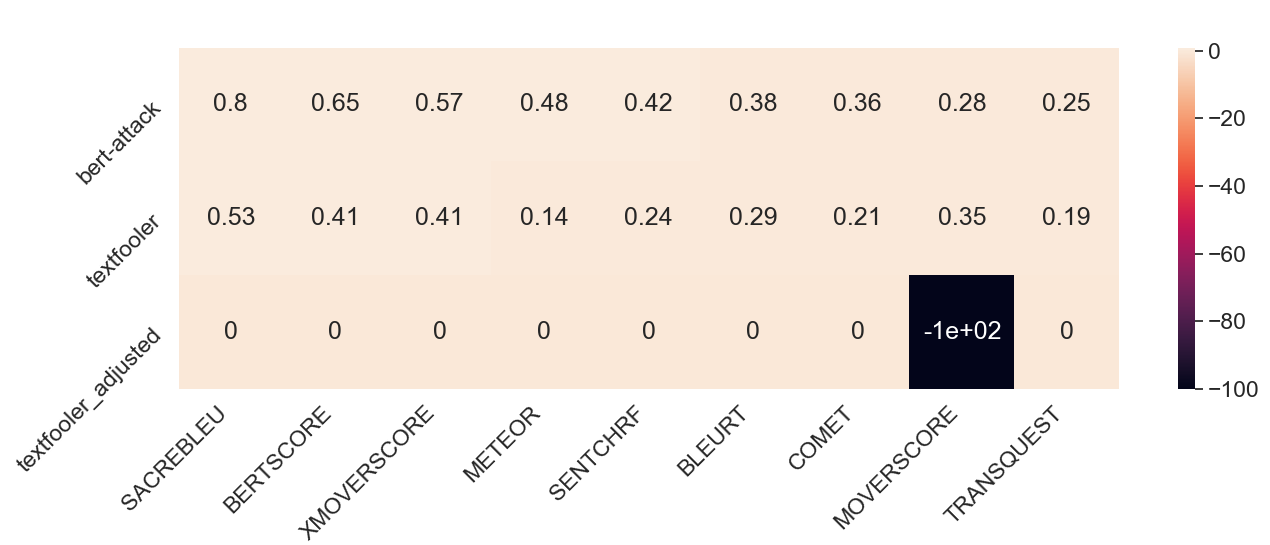}
    \includegraphics[
    scale=0.4]{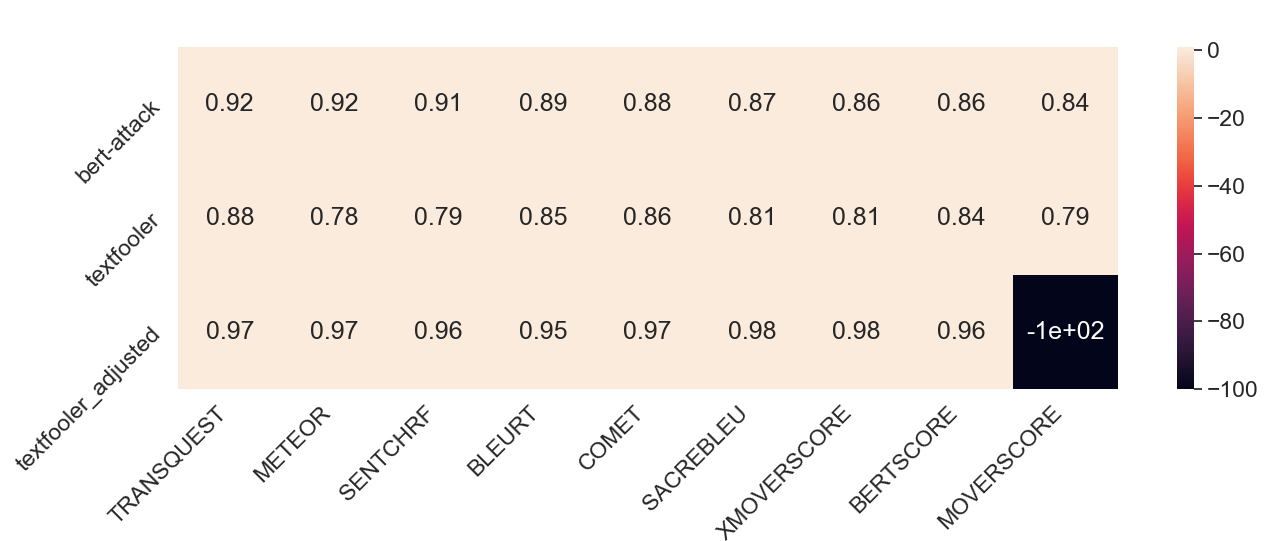}
    \caption{Top: Perturbation rate of adversarial attacks on 440 de-en samples from WMT19. 
    Middle: Avg.\ grammatical error introduced by adversarial attacks on 440 de-en samples from WMT19. 
    Bottom: Average sentence similarity of the original and perturbed hypothesis on 440 de-en samples from WMT19. The black tiles mean that no successful sample was found by the attack model for the corresponding metric. 
    }
    \label{fig:att-de-en-wmt19-pert}
 \end{figure}

In Figure \ref{fig:att-de-en-wmt19-pert}, 
we 
further show the respective perturbation rates, the number of new grammatical errors introduced, 
and the sentence similarity before and after perturbation.  
These statistics are defined in Table \ref{table:attackProperties}. 
The perturbation rates show that the BERT-Attack and TextFooler need fewer  perturbations to attack the supervised metrics, more perturbations for soft matching metrics like XMoverScore and MoverScore and the most perturbations for hard metrics. 
The pattern for TextFooler-Adjusted is again slightly different.  
BERT-Attack on average introduces more grammatical errors than TextFooler, and TextFooler-Adjusted makes fewest grammatical errors. TextFooler-Adjusted also produces most similar hypotheses (measured via sentence similarity) and TextFooler produces least similar hypotheses. 
To sum up, according to our automatic evaluation, BERT-Attack and TextFooler are highly successful in attacking supervised metrics, but they introduce more grammar errors and are less faithful to the original hypotheses than TextFooler-Adjusted. The latter has low success rates. 


\begin{table}[!htb]
    \centering
   \begin{tabular}{ll}
         \toprule
          Success Rate & \makecell[l]{Rate of successfully attacked samples\\ over all samples \citep[e.g.][]{tsai-2019}. }\\
          \midrule
          Perturbation Rate & \makecell[l]{Average number of perturbed words\\ over all words per sentence. \citep[e.g.][]{morris-2020-textattack}. } \\
          \midrule
          Rate of introduces gram. Errors & \makecell[l]{Average number of new grammatical errors \\introduced per sample, as measured \\by Language Tool\\ \citep{naber-2003,morris-2020-reeval}} \\
          \midrule
          Sentence Similarity & \makecell[l]{Average sentence similarity between\\ original and perturbed sample\\ using the SBERT toolkit\\ \citep{reimers-2019}. } \\
          \bottomrule
    \end{tabular}
    \caption{Quantitative metrics to measure the quality of adversarial attacks.}
    \label{table:attackProperties}
\end{table}

\paragraph{Human Evaluation}
To investigate 
the validity of our above results, 
we also evaluate using humans. To do so, we first split the successful attack samples into attacks to reference-based metrics and attacks to reference-free metrics. Afterwards, we sort both by sentence similarity as the first key and the number of introduced grammatical errors as the second key (i.e. two of the indicators used for metric robustness in the last paragraph). To check how effective this ordering is to identify how meaning-preserving a sample is, we randomly select samples from the top 10\% (best) and worst 10\% (worst) for the human evaluation (below, we refer to this as \textit{pre-selection}). Three co-authors of this work annotated 40 common instances of attacks, and each one of them annotated another 40 individual instances (one annotated 80 individual instances). 
Two of those three co-authors (which were bilingual) in addition annotated 40 instances in the reference-free scenario. 
In total, we thus have annotated 240 instances.

\begin{figure}[!tbp]
  \begin{subfigure}[b]{0.49\textwidth}
    \includegraphics[width=\textwidth]{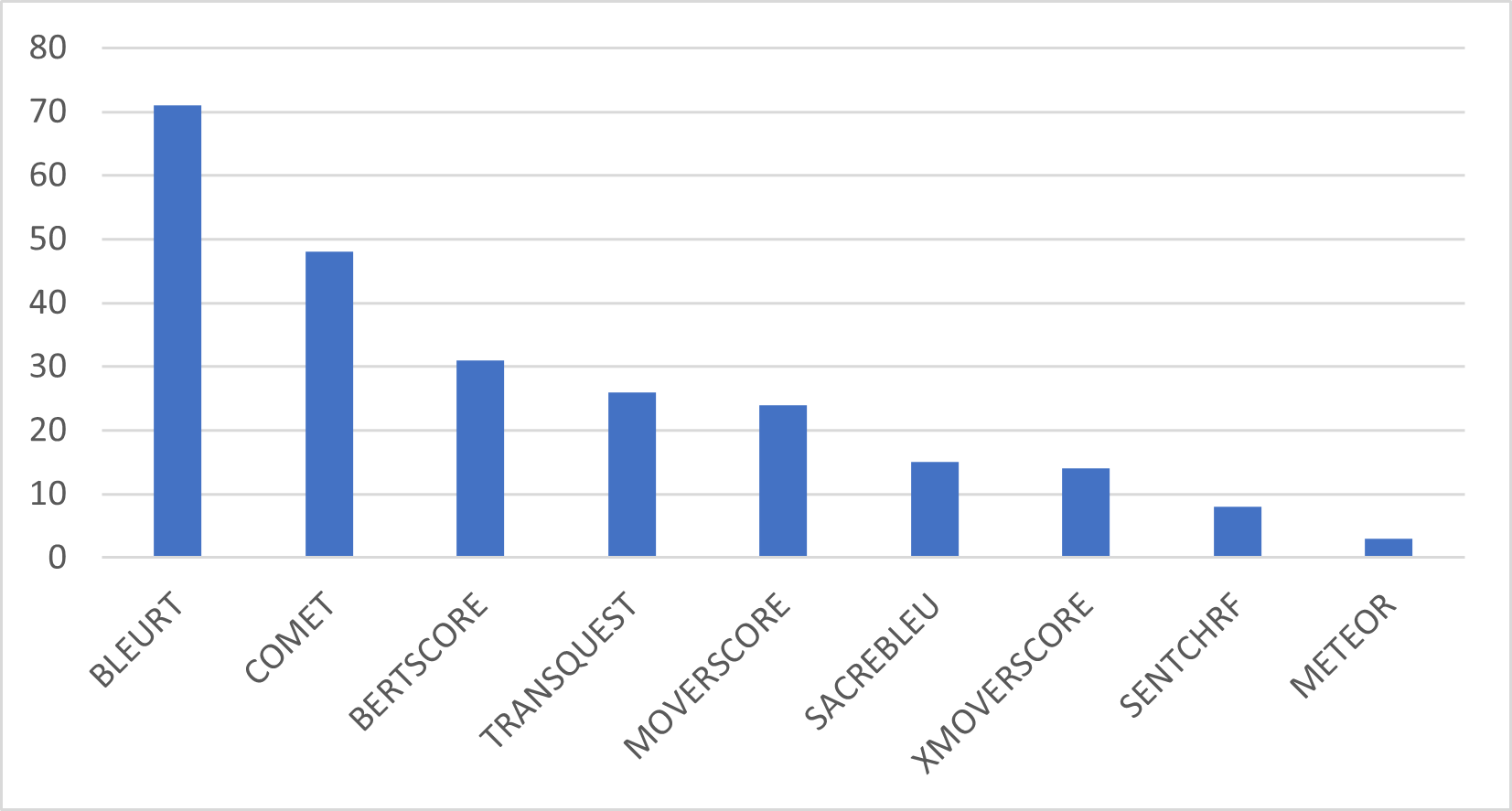}
    \caption{Samples per Metric}
    \label{fig:SamplesPerMetric}
  \end{subfigure}
  \begin{subfigure}[b]{0.49\textwidth}
    \includegraphics[width=\textwidth]{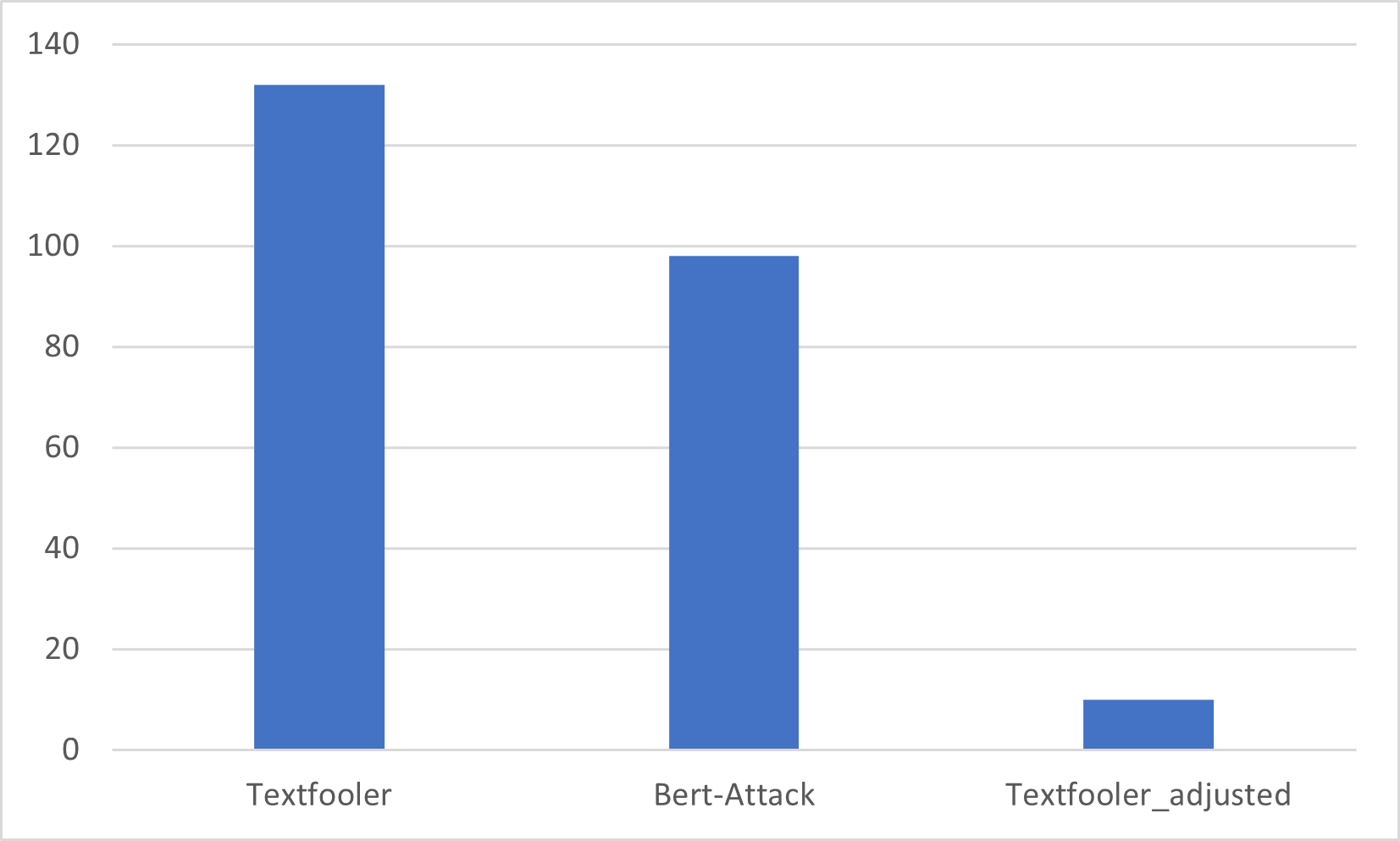}
   \caption{Samples per Attack}
    \label{fig:SamplesPerAttack}
  \end{subfigure}
  \caption{Sample distribution per metric and per attack.}
    \label{fig:DistributionAcrossDimensions}
\end{figure}

Figure \ref{fig:DistributionAcrossDimensions} shows the distribution of the 240 annotated instances across the dimensions (a) \textit{Metric} and (b) \textit{Attack Model}.  
Concerning \textit{Metric}, BLEURT is the most frequent metric appearing in our samples, METEOR the least frequent. Concerning \textit{Attack Model}, there are fewest examples from Textfooler-Adjusted, as it has lowest success rate. Further, we note that we annotated 115 of the \textit{best} and 125 of the \textit{worst} samples according to the ordering. 

The annotation scheme was whether an attacked sentence preserved the adequacy of the translation (label $=$ 0), made it worse (label $=$ 1) or considerably worse (label $=$ 2). 
Prior to annotation, there was no communication among annotators, e.g., no guidelines were established or how to deal with particular cases. 

Selected examples are shown in Table \ref{table:exemplary_human_labels}. 
Table \ref{table:human_results} shows Cohen's kappa between 3 annotators on the common set of annotated instances, Cohen's kappa for a set of annotations 
conducted twice by one of the annotators 
and statistics on the distribution of annotations. 
Inter-annotator agreement is low, with kappa slightly above 0.3. However, when only the decision `same' (label $=$ 0) vs.\ `different'  (label $=$ 1,2) is made, the agreement is acceptable among all annotators (0.64). 
A reason may be that it is often difficult to decide whether a sentence in which only one or two words are changed (as the attacks typically do) would count as `lower' or `much lower' adequacy. This may be especially difficult to judge when one (often crucial) word is changed in a long sentence.  

Out of 40 common examples, only one is labeled as `same' by all annotators; each annotator individually only labels 3 out 40 of instances as `same'.  
Across all samples, the mean of annotation labels is about 1.2 out of 2. 
On average across all annotators, about 9\% of adversarial attacks are annotated as preserving the adequacy of the original hypothesis. 
Interestingly, the supervised metric BLEURT is most frequently involved in these situations (it also occurs most frequently in our data), followed by the hard metrics SACREBLEU and SENTCHR (which occur much less frequently).

\begin{table}[!htb]
    \centering
    \begin{tabular}{cl}
         \toprule
         REF & It was a huge step in a cool season. \\
         HYP & "This was a huge step for a cool season.\\
         HYP* & "This was a \attacked{massive} step for a cool season.\\ 
         \midrule
         SRC & "Dafür ist es einfach zu früh", sagt Axel Büring.\\
         HYP & "It is simply too early for that," says Axel Büring.\\
         HYP* & "It is simply too early for that," \attacked{states} Axel Büring.\\
         \midrule
         REF & Many participants trained several hours a day. \\
         HYP & Many participants practiced several hours a day.\\
         HYP* & Many participants practiced several \attacked{calendar} a \attacked{dag}.\\ 
         \bottomrule
    \end{tabular}
    \caption{Exemplary human grades for sentences from WMT19. Top two cases: All annotators labeled `same' (involved metric: SentCHR and XMoverScore, respectively). Bottom: All annotators labeled `much worse'.}
    \label{table:exemplary_human_labels}
\end{table}

 \begin{table}[!htb]
    \centering
   \begin{tabular}{ll}
         \toprule
         Kappa, inter, reference-based (3 Annotators, 40 Samples), fine & 0.369, 0.304, 0.354 \\
         Kappa, inter, reference-based (3 Annotators, 40 Samples), coarse & 0.639, 0.639, 0.639 \\
         Kappa, inter, reference-free (2 Annotators, 40 Samples), fine & 0.323 \\
         Kappa, inter, reference-free (2 Annotators, 40 Samples), coarse & 0.642 \\
         \midrule
         Kappa, intra, reference-based (1 Annotator, 40 Samples), fine & 0.616 \\
         Kappa, intra, reference-based (1 Annotator, 40 Samples), coarse & 0.875 \\
         \midrule
         Mean human class & 1.25\\
         Avg.\ Proportion labeled `same' & 9.01\% \\
         \bottomrule
    \end{tabular}
    \caption{Results of the human evaluation, inter- and intra-annotator kappa agreement and average annotation. }
    \label{table:human_results}
\end{table}

%

\begin{figure}
  \centering
    \includegraphics[width=\textwidth]{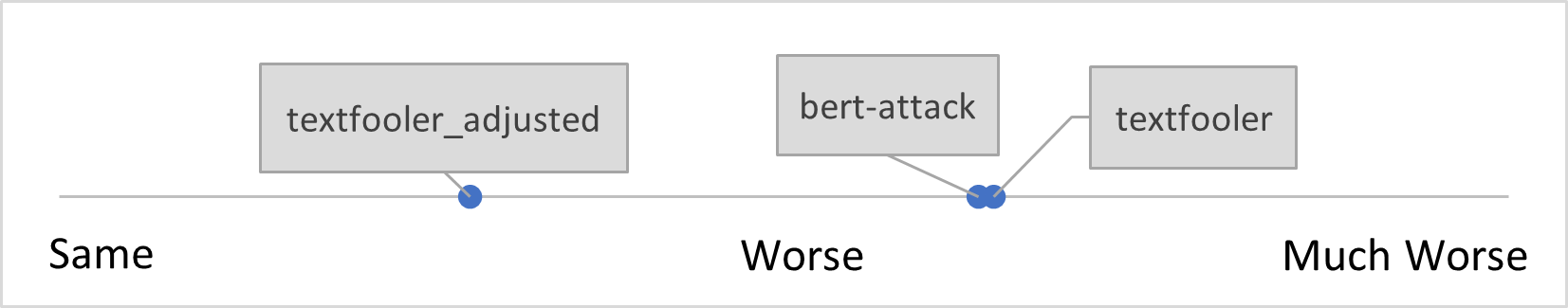}
    \caption{Mean per attack model. We assign the human labels to ``Same''=0, ``"Worse''=1 and ``Much Worse''=2.}
    \label{fig:meanPerAttackType}
 \end{figure}

\begin{figure}
  \centering
    \includegraphics[width=\textwidth]{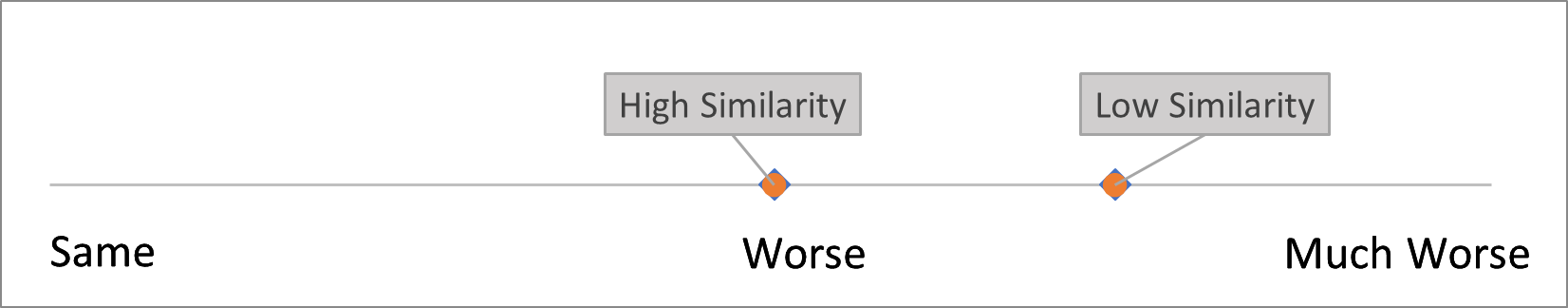}
    \caption{Mean per pre-selection. We assign the human labels to ``Same''=0, ``"Worse''=1 and ``Much Worse''=2.}
    \label{fig:meanPerPreselection}
 \end{figure}

\begin{figure}
  \centering
    \includegraphics[width=\textwidth]{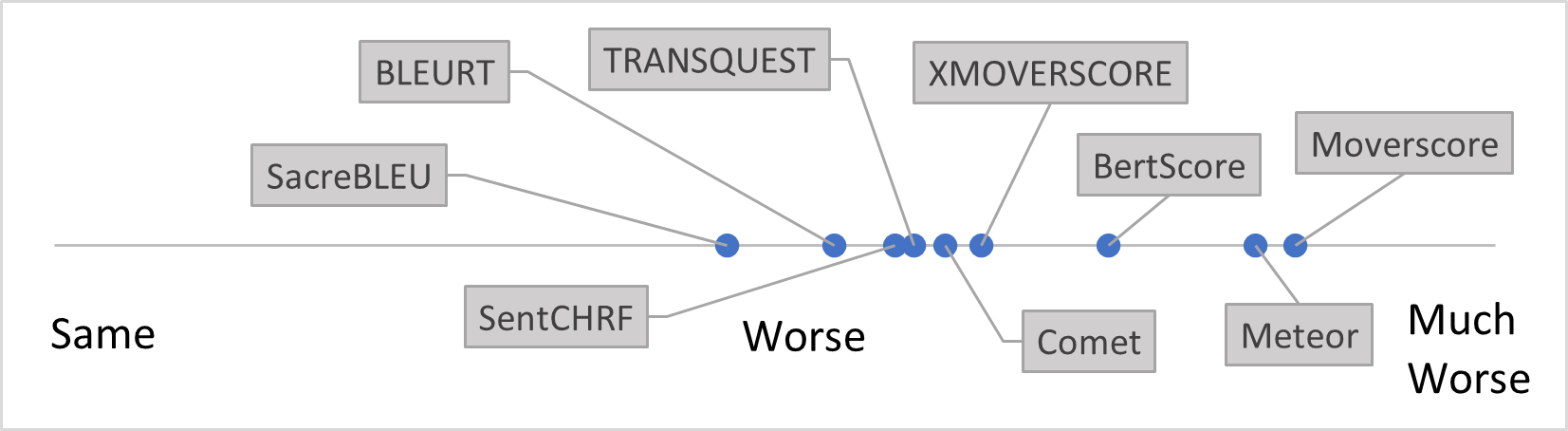}
    \caption{Mean per metric. We assign the human labels to ``Same''=0, ``"Worse''=1 and ``Much Worse''=2.}
    \label{fig:meanPerMetric}
 \end{figure}

Figure \ref{fig:meanPerAttackType} shows that TextFooler-Adjusted beats the other two attacks in human evaluation. 
For this reason, there are also fewer successes with this attack model. 
Our evaluation shows that the permutations of Bert-Attack and Textfooler not only introduce small errors, but often change the meaning of a hypothesis completely. 
Typical cases include changes of referents/entities (``In Spain'' vs.\ ``\attacked{Across Castellano}''), a complete destruction of the sense of sentences (``It is simply too early for that'' vs.\ ``It is simply too \attacked{after} for that'') or the introduction of small (grammar) errors (``After the dissolution of the band'' vs.\ ``\attacked{into} the dissolution of the band''). 
Figure 
\ref{fig:meanPerPreselection} shows that pre-selecting based on semantic similarity of another model can improve the attack quality. This is reasonable, as it can be seen as filtering for fulfilling a constraint after an attack has been applied. Figure \ref{fig:meanPerMetric} shows how reasonable the attacks on different metrics were according to the annotators. We can see that SacreBLEU (SentenceBleu) could be fooled with better attacks than other metrics; conversely, when MoverScore is fooled, the adequacy of the attacked hypothesis is lowest, indicating that the metric reacted adequately to the attack. 


We conclude 
that automatic adversarial attacks on evaluation metrics are apparently not feasible using the selected established adversarial attack models. While tools like BERT-Attack and TextFooler 
may to some degree be useful for simpler text classification tasks, 
 we find that less than 10\% of attacks on evaluation metrics are legitimate -- i.e., label-preserving -- in our human assessment (more restrictive tools such as TextFooler-Adjusted are more suited, but have much lower success rates). 
 A reason may be that evaluation metrics may be considered a harder task than classification; for example, changing an entity in a translation (``Bill is rich'' vs.\ ``Tom is rich'') may 
 not alter its sentiment, subjectivity, etc., but certainly its adequacy. 
 In other words, for MT evaluation, the notion of label-preserving 
 coincides with the notion of meaning-preserving.  
 To preserve meaning seems beyond the scope of many current adversarial techniques, however. 
 We also point out the simple design scheme of many current adversarial attack models, which aim to keep the sentence structure but only rewrite one or two important words. More clever adversarial attacks implemented via humans, such as 
 retaining the semantics but aggressively changing the surface level of the hypothesis \citep{kaster-2021} or carefully designed check lists \citep{sai-2021}, can still possibly expose important weak spots of metrics. 
%

\subsection{Hard Reference Free Metrics for the Eval4NLP Shared Task}\label{sec:hard}

\paragraph{Hard Reference Free Metrics} We provide results for four additional pairs of metric + explainability techniques, which we will refer to as \textit{hard reference-free metrics}. Hard metrics like BLEU and METEOR are interpretable (which makes them attractive for the Shared Task) and reference-based. 
In order to 
explore how well the word-level scores extracted from these metrics perform, 
we obtain pseudo references using (1) M2M100 \citep{fan-2020}\footnote{We use the library EasyNMT by Nils Reimers, \url{https://github.com/UKPLab/EasyNMT}} 
and (2) word by word (wbw) translations using a static dictionary, obtained with Google Translate.\footnote{\url{https://translate.google.com/}}  
Further, we use SHAP \citep{lundberg-2017} to extract word-level scores from BLEU and METEOR.\footnote{As BLEU and METEOR are interpretable, word-level scores could also be computed from them directly. We leave this to future work.} 

Table \ref{tab:HardRefFree} shows the results on the test sets compared with the baselines of the Shared Task\footnote{In contrast to the setting of the shared task, we only present the word-level explanation of the hypothesis, as the explanation of the source cannot immediately be retrieved from the setup. An explanation of the source might be retrieved by translating the hypothesis into the source language and applying the hard metric there.}. The results are unexpected, as the hard reference-free metrics outperform the baselines on the whole concerning word-level explanation plausibility. 
Even with word-by-word translations, they 
often outperform the Shared Task baselines. 
On average across all languages, the translated METEOR with SHAP beats Transquest with LIME by $0.21$ difference in AUC, $0.11$ difference in AP and $0.1$ difference in RtopK.\footnote{In the Shared Task, the explanations achieve rank $6.67$ for et-en, $6.33$ for ro-en, $4.67$ for ru-de and rank $7$ de-zh.} 
These findings suggest that due to the interpretability of METEOR and BLEU, SHAP is able to extract better feature importance scores than from e.g. XMoverScore. This is a counterintuitive finding, as BLEU and METEOR have been show to be inferior to soft metrics on a sentence-level. A possible reason is that the explainability techniques of the baselines have difficulties to explain the more complex models. 
This shows, however, that there is not always a relation between explanation and model output, a property that is ignored when only evaluating for plausibility. Hence, an additional evaluation for faithfulness (see Section \ref{EvalOfExplanations}) could prove beneficial for evaluations with the goal of increased model understanding. 
In Table \ref{tab:HardRefFree}, we also do not observe a correspondence between sentence- and word-level scores; the latter are  better for our novel baselines, but the sentence-level scores are substantially higher for the trained Transquest metric for the language pairs involving English (on which the metrics have been trained), further highlighting the need for faithfulness evaluation. 


 \begin{table}
\centering
\begin{tabular}{|ll|ccc|c|} \toprule
     & & \multicolumn{3}{c|}{Hypothesis} & \multicolumn{1}{c|}{} \\
    \textbf{Lang.} & \textbf{System} & AUC & AP & RtopK & Pearson\\\midrule
et-en & TranslationBLEU$_{\text{full}}$ + SHAP & \textbf{0.754} & 0.541 & 0.418 & 0.482 \\
& TranslationMETEOR$_{\text{full}}$ + SHAP & \textbf{0.754} & \textbf{0.562} & \textbf{0.444} & 0.581 \\
& TranslationMETEOR$_{\text{wbw}}$ + SHAP & 0.670 & 0.445 & 0.323 & 0.318 \\ \hdashline
& Transquest+LIME (Baseline) & 0.62 & 0.54 & 0.43 & \textbf{0.77}\\
\midrule
ro-en & TranslationBLEU$_{\text{full}}$ + SHAP & \textbf{0.831} & 0.592 & 0.446 & 0.562 \\
& TranslationMETEOR$_{\text{full}}$ + SHAP & 0.826 & \textbf{0.603} & \textbf{0.458} & 0.701 \\
& TranslationMETEOR$_{\text{wbw}}$ + SHAP & 0.752 & 0.493 & 0.342 & 0.397 \\ \hdashline
& Transquest+LIME (Baseline) & 0.63 & 0.52 & 0.42  & \textbf{0.90}\\\midrule
ru-de & TranslationBLEU$_{\text{full}}$ + SHAP & 0.795 & 0.525 & 0.420 & 0.490 \\
& TranslationMETEOR$_{\text{full}}$ + SHAP & \textbf{0.806} & \textbf{0.546} & \textbf{0.435} & \textbf{0.597} \\
& TranslationMETEOR$_{\text{wbw}}$ + SHAP & 0.679 & 0.399 & 0.288 & 0.273 \\ \hdashline
& Transquest+LIME (Baseline) & 0.40 & 0.26 & 0.16 & 0.50\\\midrule
de-zh & TranslationBLEU$_{\text{full}}$ + SHAP & 0.633 & 0.391 & 0.261 & 0.275 \\
& TranslationMETEOR$_{\text{full}}$ + SHAP & \textbf{0.647} & \textbf{0.420} & \textbf{0.294} & 0.376\\
& TranslationMETEOR$_{\text{wbw}}$ + SHAP & \textbf{0.617} & \textbf{0.390} & \textbf{0.273} & 0.222\\ \hdashline
& Transquest+LIME (Baseline) & 0.46 & 0.27 & 0.14 & \textbf{0.34}\\
\bottomrule
\end{tabular}
\caption{Performance of ``hard reference-free metrics'' on the Eval4NLP shared task. Metric$_{\text{full}}$ indicates that a full translated sentence was used as pseudo-reference. Metric$_{\text{wbw}}$ indicates that the word by word translated source was used as pseudo-reference. Bold values indicate the best score per column and language pair.}
\label{tab:HardRefFree}
\end{table}

\subsection{Rigorous system comparison}\label{sec:bt}
In the Eval4NLP shared task, systems are ranked according to their global independent statistics,  
e.g., \textit{mean} AUC scores of different systems over a common set of test instances. 
However, aggregation mechanisms such as the \textit{mean} ignores which system beats others over individual instances, and thus 
may lead to false conclusions. 
For instance, \citet{bt-2021} 
illustrate that a system that is worse than another on all instances but one (an outlier) might be falsely declared the winner according to \emph{independent} statistics such as the \textit{mean} or \textit{median}. 
As remedy, they suggest to use the \textit{BT} (Bradley-Terry) model \citep{bt_one}, 
which performs \emph{paired} evaluation, to conduct rigorous comparison for competing systems. \textit{BT} leverages instance-level pairing of metric scores from different systems\footnote{\textit{BT} is a statistical model used to maximize the likelihood of \wei{the given number of instances where one system is better than  another},
aiming to estimate unknown, inherent strengths of systems. In the case of two systems, the strength of system A is equivalent to the percent of instances of A better than system B.}, and assumes that a winning system should beat others over the majority of instances. 
In the concrete case -- the shared task -- this would mean that a system could have very high AUC scores on few instances, which inflate its \textit{mean} AUC, but otherwise performs worse in the majority of instances.

\begin{table}
\centering
\begin{tabular}{ l | c c c c}
\toprule
& de-zh & ro-en & et-en & ru-de\\
\midrule
source.ap-scores   & $9\%$ & $14\%$ & $7\%$ & $3\%$\\
source.auc-scores       & $3\%$ & $7\%$ & $9\%$ & $0\%$\\
source.rec-topk-scores     & $38\%$ & $20\%$ & $23\%$ & $14\%$\\
target.ap-scores     & $4\%$ & $10\%$ & $2\%$ & $2\%$\\
target.auc-scores     & $2\%$ & $8\%$ & $4\%$ & $0\%$\\
target.rec-topk-scores     & $31\%$ & $19\%$ & $13\%$ & $13\%$\\
\bottomrule
\end{tabular}
\caption{Disagreement of system rankings between \textit{mean} and \textit{BT} across six evaluation metrics and four language pairs. Each cell shows the percent of system pairs ordered differently by \textit{mean} and \textit{BT} according to the recalled version of Kendall's $\tau$ supported on $[0, 1]$. Higher scores indicate higher disagreement.}
\label{tab:global_results}
\end{table}     

\begin{figure*}
\centering
\includegraphics[width=0.7\textwidth]{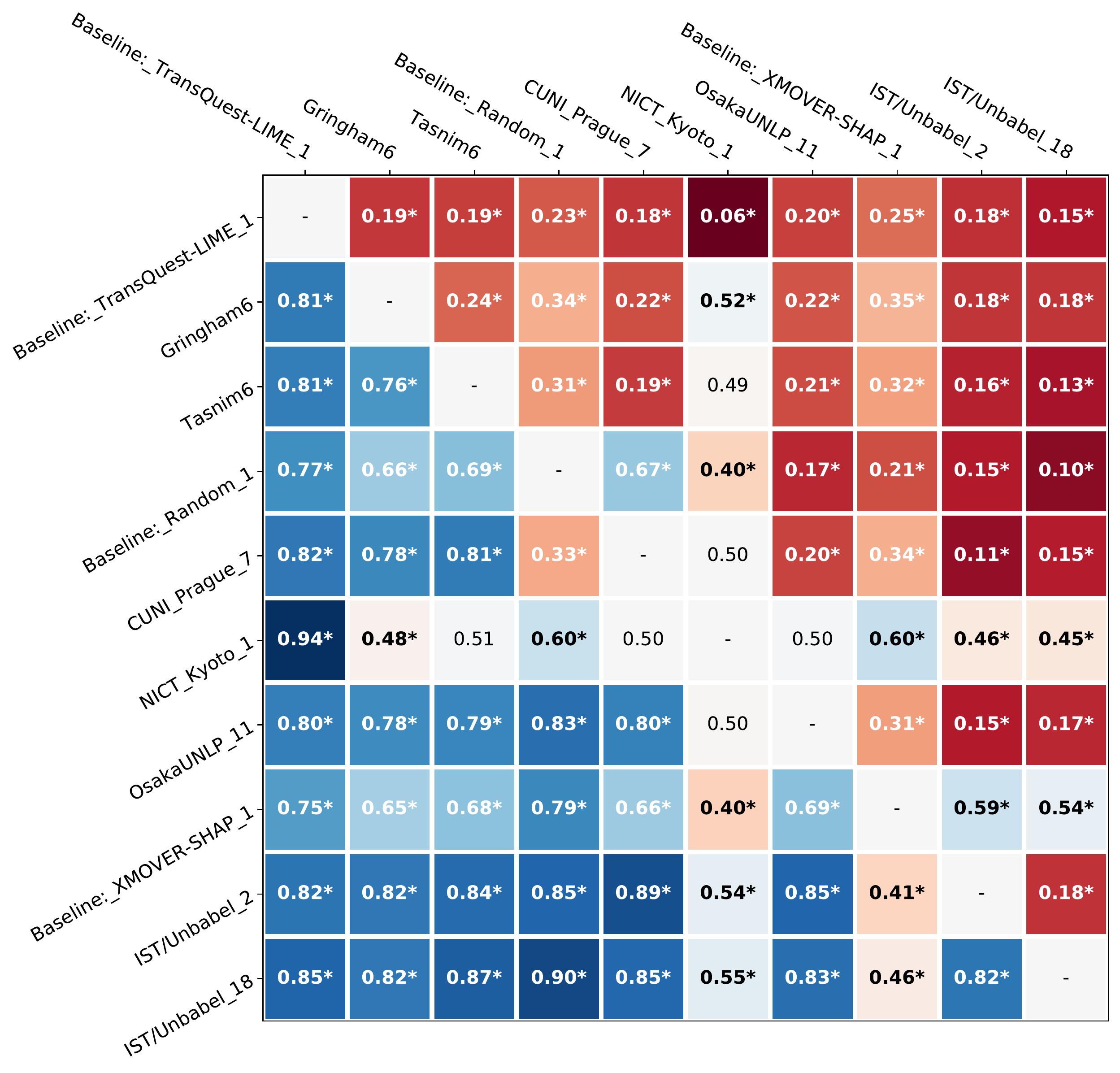}
\includegraphics[width=0.7\textwidth]
{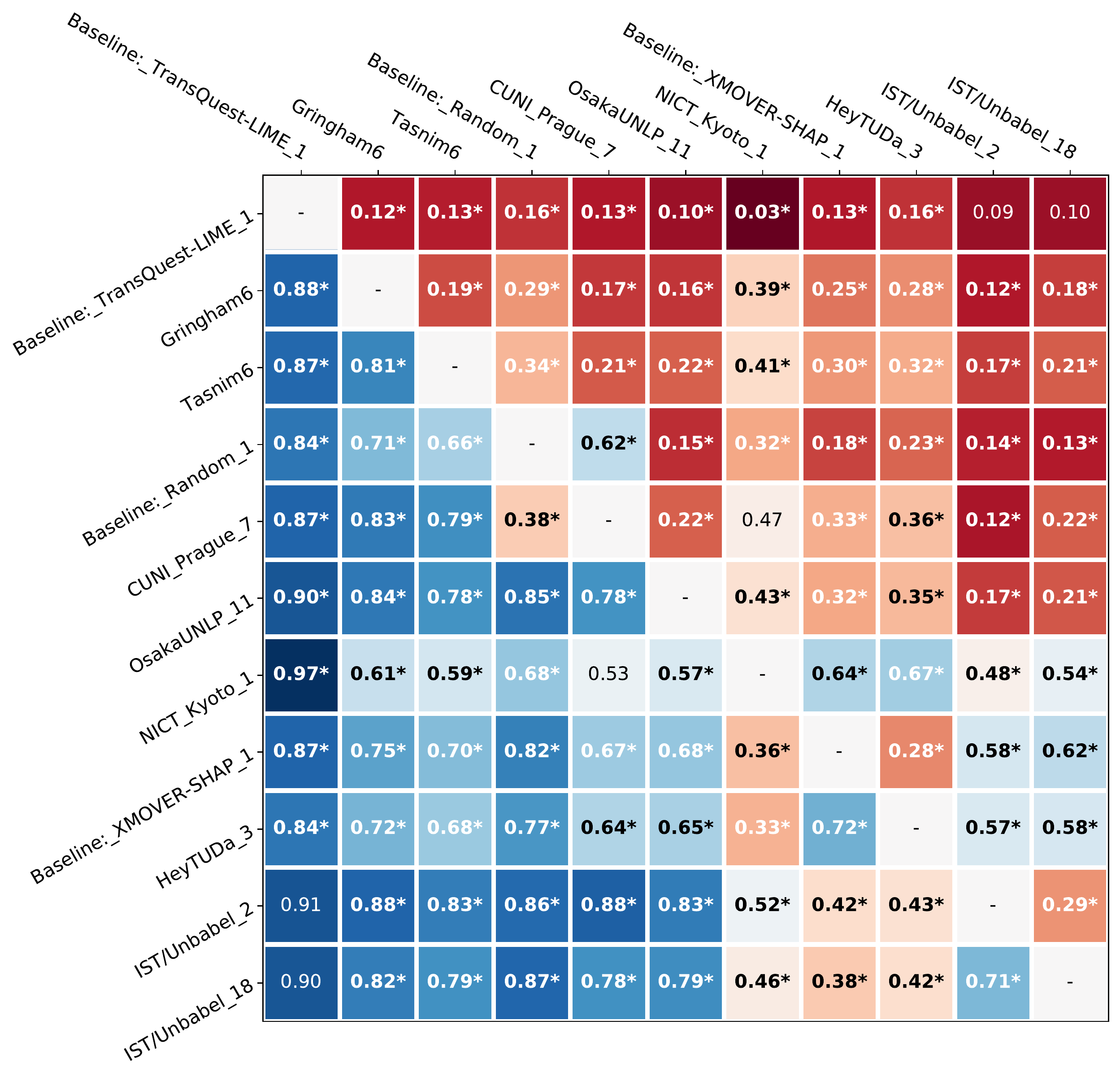}
\caption{Results of pairwise comparison according to (top) ``source.rec-topk-scores'' and (bottom) ``target.rec-topk-scores'' over system pairs for German-Chinese. Each cell denotes the percent of the 
instances in which one system (in rows) beats another (in columns). We mark cells 
for which system pairs have significant differences according to Sign test with `*'. Systems have been ranked reversely by \textit{BT}, e.g., systems in final rows are declared the best. \textit{mean} declares Kyoto-1 as the best in both (top) and (bottom) settings.}
\label{fig:bt}
\end{figure*}



We analyze whether 
\textit{mean} and \textit{BT} yield similar results on the shared task.
First, we quantify the disagreement between \textit{mean} and \textit{BT}. Table \ref{tab:global_results} shows that \textit{mean} and \textit{BT} often disagree on the ranking of systems, especially for ``source.rec-topk-scores'' and ``target.rec-topk-scores''. 
\wei{This might undermine the reliability of these recall-based metrics, as they are very sensitive to the aggregation scheme (\textit{BT} vs.\ \textit{mean}), 
unlike `ap-scores' and `auc-scores' that consider both precision and recall.}

We then provide justifications to understand the judgments of \textit{BT} and \textit{mean} on German-Chinese as use case (see Figure \ref{fig:bt}). 
We find that \textit{BT} and \textit{mean} both may yield wrong judgments as to which system is the state-of-the-art.
We illustrate this below:

\begin{itemize}
\item 
\textit{mean} might be wrong (Fig.~\ref{fig:bt}, top): Considering plausibility of explanations on source sentences, \textit{mean} declares Kyoto-1 as the best system; however, it significantly outperforms merely 3 out of 9 systems 
according to pairwise comparison. This indicates that \textit{MEAN} results are very likely wrong. In contrast, \textit{BT} chooses Unbabel-18 according to that it wins in 8 out of 9 cases. 

\item \textit{BT} might be wrong (Fig.~\ref{fig:bt}, bottom): Considering plausibility of explanations on target sentences, \textit{BT} declares Unbabel-18 as the best, as it beats 7 out of 10 systems with clear wins. On the other hand, Kyoto-1 (ranked 5th according to \textit{BT}) wins in 9 out of 10 systems, and it also beats Unbabel-18. This means Kyoto-1 might be the winner, but that \textit{BT} nevertheless favors Unbabel-18 most as \textit{BT} considers
\wei{the number of instances of one system superior to another}.
Concretely, though Kyoto-1 beats the greatest number of systems, it outperforms these systems \wei{on slightly over half of instances}, which reflects the weak \wei{strength} from a \textit{BT} perspective. In contrast, Unbabel-18 wins globally on the greatest number of instance-level pairs assembled across all systems. We depict this issue of \textit{BT} as the inconsistency between global and local judgments, i.e., that one locally beats another in the case of two systems, but the judgment of system superiority may change in the global view when involving more systems in the comparison. As \citet{bt-2021} state, the `inconsistency' can hardly be addressed according to the Arrow's impossibility theorem \cite{arrow1950difficulty}.

\end{itemize}

Our analysis shows how subtle the evaluation of systems 
which (in the case of the shared task) explain MT metrics can be and that there may be no clearly best explainability model, as none of the systems beats all other systems according to pairwise comparison. We recommend future such evaluations to consider multiple aggregation schemes (including \textit{mean} and \textit{BT}) for a more fine-grained assessment. 


\section{Future Work} \label{sec:future}
In this section, we lay out ideas for future explainability approaches of MT metrics.

\subsection{Text generation as explainability for text generation metrics} 
\label{InverseMetrics}

Providing explanations in text form \citep{camburu2018snli, kumar-talukdar-2020-nile} may be particularly attractive to human users. For MT systems, this would mean that a metric not only outputs one or several scores, but also generates a textual explanation for the metric score. 
Below, we provide a vision of concrete text generation approaches as explanation for MT 
but we also point out that we could even more radically envision a new class of holistic generalized MT systems that output translations, scores, and human-understandable explanations. 

\paragraph{Inverse Metrics}
Adversarial attacks 
produce examples that lie as close to the original input as possible. The property which makes them interesting for local explanations is that the change can still be perceived, which yields insights for the user or developer. 

However, an explanation 
does not necessarily have to be close to the original sentence. To that end, we introduce \textit{inverse metrics}, which can be interpreted as a special form of adversarial attacks or counterfactual examples.\footnote{A similar approach can be found in Computer Vision, where CNNs are explained by inverting their computation \citep{mahendran-2014}.} 
We define an inverse metric METRIC$^{-1}$ as follows:
\begin{align}
\text{METRIC}^{-1}(\text{s},\text{score})=\text{HYP*}\:\iff\:\text{METRIC}(\text{ s},\text{HYP*})=\text{score}
\end{align}
where $s$ is the source or reference.

In other words, the inverse metric returns a hypothesis for which the metric will return a given score. 
For example, if a good reference-free machine translation evaluation metric assigns a score of 1 to a perfect translation of a source, the inverse metric of the source and the score 1 should ideally return a perfect translation as well. 
If the lowest possible score is 0, the inverse metric of the source and the score 0 should return a translation that is as irrelevant as possible (as measured by the metric).
Note that the output of an inverse metric could be interpreted as a targeted, unconstrained adversarial attack on metrics with numeric outputs, or, depending on the definition, as a counterfactual example. 
When the target score is equal or similar to the original score, the output can be viewed as a prototypical example for the score. 

\begin{figure}
    \centering
    \includegraphics[width=0.7\textwidth]{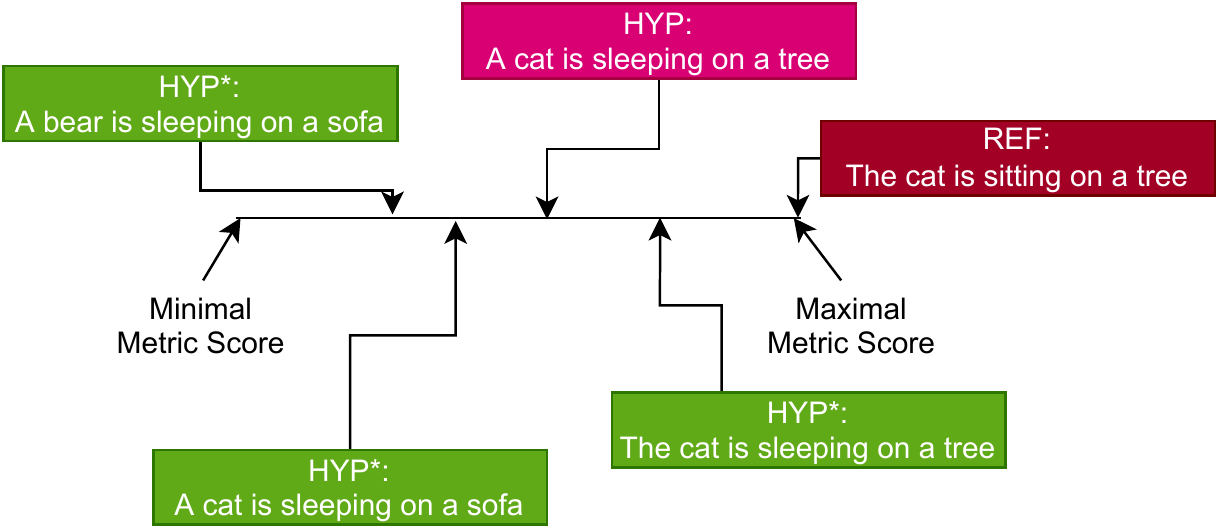}
    \caption{Inverse Metrics: Hypothetical examples around a given hypothesis.} \label{counterfactualExplanation}
\end{figure}

Figure \ref{counterfactualExplanation} gives a hypothetical example of how such examples can provide insights into a model's local behavior. By sampling other hypotheses around the original score, it could be shown what the metric indicates as better or worse than the original hypothesis. In case of metrics with a good performance, this could be used to improve translations. In case of weak metrics, it could show the failure cases of the metric, e.g.\ if the inverse metric finds hypotheses that are grammatically worse but get higher scores. 

To implement inverse metrics, we used a technique that iteratively perturbs words based on mask replacement, to find an input that results in a score as close to the desired target score as possible (see Appendix \ref{sec:InverseTech} for details). 
Table \ref{table:inversealgres} shows example neighborhoods we queried for BLEU \citep{papineni-2002} and BLEURT \citep{sellam-2020} (we use BLEURT-base-128). I.e. we use an interpretable metric and a black box metric. Based on these results, we can verify that BLEU does not try to preserve semantics as an unrelated (lexical overlap) sentence such as ``My cat is at home'' receives a better score than the original hypothesis. 
Further, for BLEURT we acquire a rather interesting improvement ``My cat dates back two generations .''. While close in meaning, this is not a phrase that would be used in natural language. As the technique produces a neighborhood for a sample, it might also be integrated into other explainability techniques which require neighborhood examples for their computation (e.g. LIME). 

\begin{table}[!htb]
    \centering
    \begin{tabular}{cll}
         \toprule
         \textbf{REF} & My cat is old. &  \\
         \textbf{HYP} & My cat lives since 17 years. &  \\ \midrule\midrule
         \textbf{BLEU} &  & Orig. Score: 0.253 \\
         \midrule
         \textbf{HYP*} & The boy lives since then ? & Pert.\ Score: 0.0 \\
         \midrule
         \textbf{HYP*} & My family lives since then ? & Pert.\ Score: 0.193 \\
         \midrule
         \textbf{HYP*} & My cat is on fire ? & Pert.\ Score: 0.398 \\
         \midrule
         \textbf{HYP*} & My cat is on 17 . & Pert.\ Score: 0.427 \\
         \midrule
         \textbf{HYP*} & My cat lives since 17 . & Pert.~Score: 0.302 \\
         \midrule
         \textbf{HYP*} & My cat is at home . & Pert.\ Score: 0.427 \\
         \midrule
         \textbf{BLEURT} &  & Orig. Score: -0.514 \\
         \midrule
         \textbf{HYP*} & This cat survived about three years . & Pert.\ Score: -0.999 \\
         \midrule
         \textbf{HYP*} & My cat lived for 17 years . & Pert.\ Score: -0.594 \\
         \midrule
         \textbf{HYP*} & My cat has aged ten years . & Pert.\ Score: -0.111 \\
         \midrule
         \textbf{HYP*} & My cat dates back two generations . & Pert.\ Score: 0.115 \\
         \midrule
         \textbf{HYP*} & My cat lives since 17 years . & Pert. Score -0.514 \\
         \midrule
         \textbf{HYP*} & My cat has survived many years . & Pert. Score -0.184 \\
         \bottomrule
    \end{tabular}
    \caption{Inverse metrics: exemplary hypothesis neighbourhood generated with the algorithm in Appendix \ref{sec:InverseTech}. From top to bottom, the target scores were chosen to be $[0, 0.2, 0.4, 0.6, 0.8, 0.9]$ for BLEU and $[-1, -0.6, -0.2, 0.2, 0.6, 0.8]$ for BLEURT.}
    \label{table:inversealgres}
\end{table}



\subsection{Explainability for Discourse Metrics}

As text generation systems become better and better, more and more MT systems will expectedly operate on document-level in the future \citep{voita-etal-2019-good}, rather than on sentence-level, as is the current standard. The corresponding evaluation metrics will need to be able to take sentence-level context into account as well. This is an emergent topic in NLG evaluation, see e.g.\ \citet{jiang2021blond,zhao2022discoscore}. 
For instance, \citet{joty2017discourse} proposed DISCOTK to address discourse coherence evaluation, which compares hypothesis with reference on the level of rhetorical structures. \citet{jiang2021blond} presented BLOND, which measures the consistency of gender and verb tense between hypothesis and reference.  \citet{zhao2022discoscore} introduced DiscoScore, which compares readers’ focus of attention in hypothesis with that in reference.
Among the three, BLOND and DISCOTK are transparent metrics, both of which adopt simple statistics to measure the inconsistency on the levels of verb tense, gender and structures. 
To achieve higher quality, DiscoScore is 
based on blackbox language models, 
which makes it non-transparent. 
Even though \citet{zhao2022discoscore} provided justifications to the superiority of DiscoScore over other metrics, little is known how much the `blackbox' judgments are trustworthy---which is one of the major goals in explainable artificial intelligence \cite{arrieta2020explainable}, and also in this work. 


As for explainable high-quality document-level evaluation metrics, a range of post-hoc techniques could play a vital role for understanding non-transparent discourse metrics as extensions of what has been surveyed in this work for sentence-level metrics, e.g., providing rationales to the judgments of these metrics in the form of (i) importance distribution, viz., the probability distribution of words in hypothesis that exhibit discourse errors; (ii) simpler, transparent models such as linear regression and decision trees; (iii) generated textual explanations, etc. 

The challenge of such extensions lies in the additional complexity of explainable document-level metrics.
For example, error annotations would also need to highlight cross-sentence relations and account for divergent linguistic structure.
Also, transparent surrogate models (e.g., linear regressions) explaining the blackbox ones would need to be able to take cross-sentential context into account, which involves cross-lingual discourse phenomena such as ``coreference resolutions in source and hypothesis''.  
Adversarial examples on the document-level (e.g., wrong gender agreement, wrong reference relations) would be particularly insightful for the development of better document-level metrics.   

\subsection{Leveraging the interplay between word-level explanations and sentence-level metrics}
In the Eval4NLP shared task, explainability techniques were used to explain sentence-level metrics scores by word-level scores, yielding a plausibility evaluation. On the other hand, \citet{freitag-2021} and also a few Eval4NLP shared task participants (e.g., \citet{polak-2021}) show that word-level scores may be used to infer sentence-level scores. This is an interesting duality between word-level and sentence-level metrics, which future work may exploit. A particular appeal lies in the fact that word-level rationales may be extracted from sentence-level metrics in an unsupervised manner using the explainability techniques, giving rise to self-supervised improvement techniques \citep{Belouadi2022USCOREAE}. 

\subsection{Extrinsic Evaluation of Explanations for MT Metrics}
As discussed in Section~\ref{EvalOfExplanations}, explanations could be evaluated intrinsically (with respect to some desirable properties) and extrinsically (measured via improved outcome on downstream tasks after incorporating the explanations).
Concerning intrinsic evaluation, we have seen the 2021 Eval4NLP Shared Task \citep{fomicheva-2021} focusing on evaluating plausibility of the explanations, i.e., how sensible the relevance scores are when compared to word-level errors. 
The evaluation of generated adversarial samples in Section~\ref{sec:AdversarialAttacks} is another example of intrinsic evaluation.
However, only few existing works on explainability for MT metrics (e.g., \citet{sai-2021}) conduct extrinsic evaluation.
In other words, most related works do not check whether their explanations can truly help achieve goals discussed in Section \ref{sec:xmte}: providing further information for non-experts and non-native speakers, diagnosing and improving the metrics, increasing efficiency of annotators for word-level translation errors, etc.
Hence, it would be interesting for future work to test these goals practically. 
For instance, one may develop an annotation tool which shows explanations for MT metrics as supporting information and measure human annotators' efficiency, compared to the case where they use the system with no explanation.
Also, developing a new framework for incorporating human feedback on different types of explanations to improve the metric is another way to evaluate the explanations with respect to a downstream task (i.e., metric improvement) \citep{lertvittayakumjorn-2021}. 
Lastly, it is also possible to measure user trust in the metrics with and without the explanations so as to assess whether the explanations can boost the user trust and promote adoption of complex model-based metrics \citep{hoffman2018metrics,jacovi2021formalizing}.



\subsection{Sentence-Pair Feature Importance} 
With GMASK, \citet{chen-2021} introduce an explainability technique that provides hierarchical feature attributions on sentence pairs.
To explain, the technique forms groups of words between two input sentences and assigns each group an importance score. This approach is 
highly relevant for MT metrics, as these are based on sentence pairs in most cases. For example, when one word in a source is translated into multiple words in the hypothesis, GMASK could identify this connection and provide joint importance values. In particular, the method seeks to fulfill three core properties: (1) providing correct importance selections, (2) considering only the most relevant words, (3) masking correlated words together. GMASK learns binary masks that indicate for each embedding in the two input sentences whether it should be dropped or kept. Thereby, it tries to keep the originally predicted output while reducing the number of relevant words. Ideally, when used for explainable MT evaluation, the approach might provide outputs like the hypothetical example shown in table, where each color indicates a different group \ref{table:gmask_sample}.

\begin{table}[!htb]
    \centering
    \begin{tabular}{cl}
         \toprule
         \textbf{REF} & \textcolor{blue}{I have} an \textcolor{orange}{apple}, which is \textcolor{orange}{green} . \\
         \textbf{HYP} & \textcolor{blue}{My} \textcolor{orange}{apple} is \textcolor{orange}{green} .\\
         \bottomrule
    \end{tabular}
    \caption{Hypothetical explanation with GMASK. The color indicate words that are grouped together.}
    \label{table:gmask_sample}
\end{table}

\citet{chen-2021} 
evaluate their approach on a dataset for natural language inference (NLI) and three datasets for paraphrase identification. To evaluate faithfulness, they 
calculate an AOPC score, post-hoc accuracy, and perform a degradation test (see Section \ref{Faithfulness}) to check how well the most-relevant and least-relevant tokens predicted by GMASK influence the original outcome. 
Overall, their results indicate that their method is the most faithful compared to methods that target single sentence inputs.

GMASK has not yet been applied to evaluation metrics. A particular challenge is that metrics give continuous scores, while GMASK has hitherto been applied for classification tasks. Also, GMASK employs a preselection approach that is problematic in case of MT metrics, as it will likely drop errors in favour of correctly translated words. If these challenges are overcome we fathom that this approach could provide strong explanations.




\section{Conclusion} \label{sec:conclusion}

{
The difficulty of understanding machine learning models has implications on their real world usage. For example, it is dangerous to employ black-box systems in safety critical applications \citep{rudin-2019}. Also, they might unknowingly incorporate biases such as gender, racial, political or religious bias \citep{mehrabi-2021}. In their general data protection regulation, the European Union even requires that decisions that impact a person can be explained \citep{goodman-2017}. 
Hence, the interpretability and explainability of 
machine learning 
models forms a gateway to their broader usage. 
Further propelled by recent challenges, such as the eXplainable AI challenge by \citet{gunning-2017} or the explainable ML (xML) challenge by the \citet{fico-2018}, a large body of research considers this problem. 

}

In this work, we provide a taxonomy of goals and properties for \emph{explainable evaluation metrics}, a nascent field that may help overcome the dominance of classical low-quality evaluation metrics. 
 We also synthesize and categorize recent approaches to explainable evaluation metrics, highlighting their results and limitations. Currently, the two dominant approaches to explainability for evaluation metrics include (1) highlighting erroneous words in source and target that explain a sentence-level score and (2) manually devising adversarial attacks that expose weak spots of metrics and which can then be used to diagnose and improve metrics. The major weaknesses of the current realization of these techniques are that (1i) the error highlights do not consider the correspondence between words in source and target and (1ii) the severity of errors and (1iii) the evaluation does not consider the faithfulness of explanations. Further, (2i) adversarial attacks to evaluation metrics concurrently require human design and (2ii) automatizing this process is very difficult as we show, since adversarial attacks need to be meaning-preserving (which is harder than what current adversarial techniques aim for). We also present a vision of future approaches to explainable evaluation metrics, which should (A) help fix the problems of the above paradigms (e.g., via joint consideration of the sentence pairs involved),  (B) go beyond the named approaches and also consider textual explanations (which may be easier to comprehend for humans), 
 (C) leverage explainability techniques to unsupervisedly improve sentence-level metrics, 
 (D) target document-level metrics (which exhibit an additional layer of complexity) and (E) provide extrinsic evaluation across a range of different explanation types.   

 Our broader vision is that explainability is now a `desirable but optional' feature,
but we argue that in the future it will become 
essential, even compulsory, 
especially for evaluation metrics as a highly sensitive task 
assessing the quality (and veracity) 
of translated information content. 
Explainability builds transparency and trust
for users, eases bug-fixing and 
shortens improvement cycles 
for metric 
designers, and will be required by law/regulations
for AI systems to be applied to large-scale, high-stack domains. 
In this context, we hope our work will 
catalyze efforts on the topic of explainable evaluation metrics for machine translation and more general text generation tasks.

\bibliography{survey}

\appendix

\section{Inverse Metric Techniques}
\label{sec:InverseTech}
The approach we use for inverse metrics applies simple, greedy perturbations on a word-level. To do so, it randomly searches through mask replacements by a language model. Hence, it is similar to language-model-based adversarial attacks \citep[e.g.][]{li-2020}. Given a metric MTE, a hypothesis $h$ with tokens $h = (h_1, ..., h_n)$, a perturbed hypothesis $h^\ast$, a target score $t$, and a source (and/or reference) $s$, the method aims to minimize the following equation:
$$\text{argmin}_{h^\ast} |t - \text{MTE}(s, h^\ast)|$$
Further, following the perturbation setup in LIME \citep{ribeiro-2016}, let $m = (m_1, ..., m_n)$ be a mask for $h$ where each element indicates whether the respective token in $h$ is perturbed. Instead of $m$ being a binary vector, we choose each $m_i$ in $m$ to represent the $k$-th likeliest mask replacement based on a language model. E.g., a mask $m = (0, 2, 1)$ indicates that the first token is not perturbed, the second token is perturbed with the 2nd most likely word and the third token is perturbed with the most likely word. We apply the perturbations one by one in a random order. The algorithm searches for the mask that produces the best hypothesis solving the minimization problem above. We search masks for $x$ iterations. Further, we keep a list of $b$ masks per iteration, which is initialized with masks without perturbation (all zero). In each iteration, for every mask in $b$, we randomly increase each mask element by $1$ with a perturbation probability of $p$. Setting $p$ small, and $b$ high should lead to a better consideration of perturbations that are close to the original hypothesis. Instead, setting a high perturbation rate and a small number of samples per iteration will lead to far examples being explored earlier. While being easy to implement, shortcomings are that the number of tokens (words or sub-words) is already set, making some options unreachable. Additionally, the algorithm does not directly consider the source sentence.

\section{Explanation with respect REF and/or HYP}
\label{sec:fixed_parts}
\begin{figure}[]
    \centering
    \includegraphics[width=0.9\textwidth]{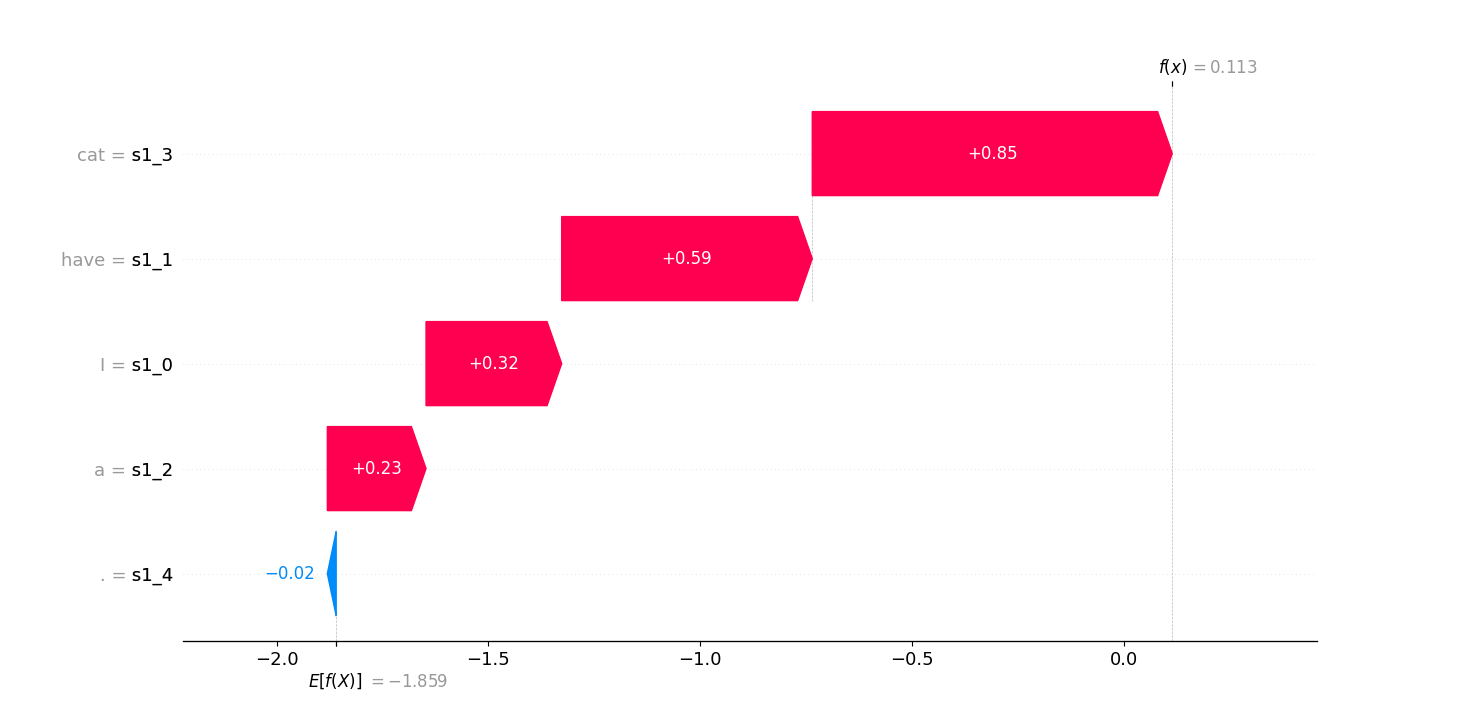}
  \caption{Explanation with respect to a fixed reference}
  \label{fig:ref_fixed}
\end{figure}

\begin{figure}[]
    \centering
    \includegraphics[width=0.9\textwidth]{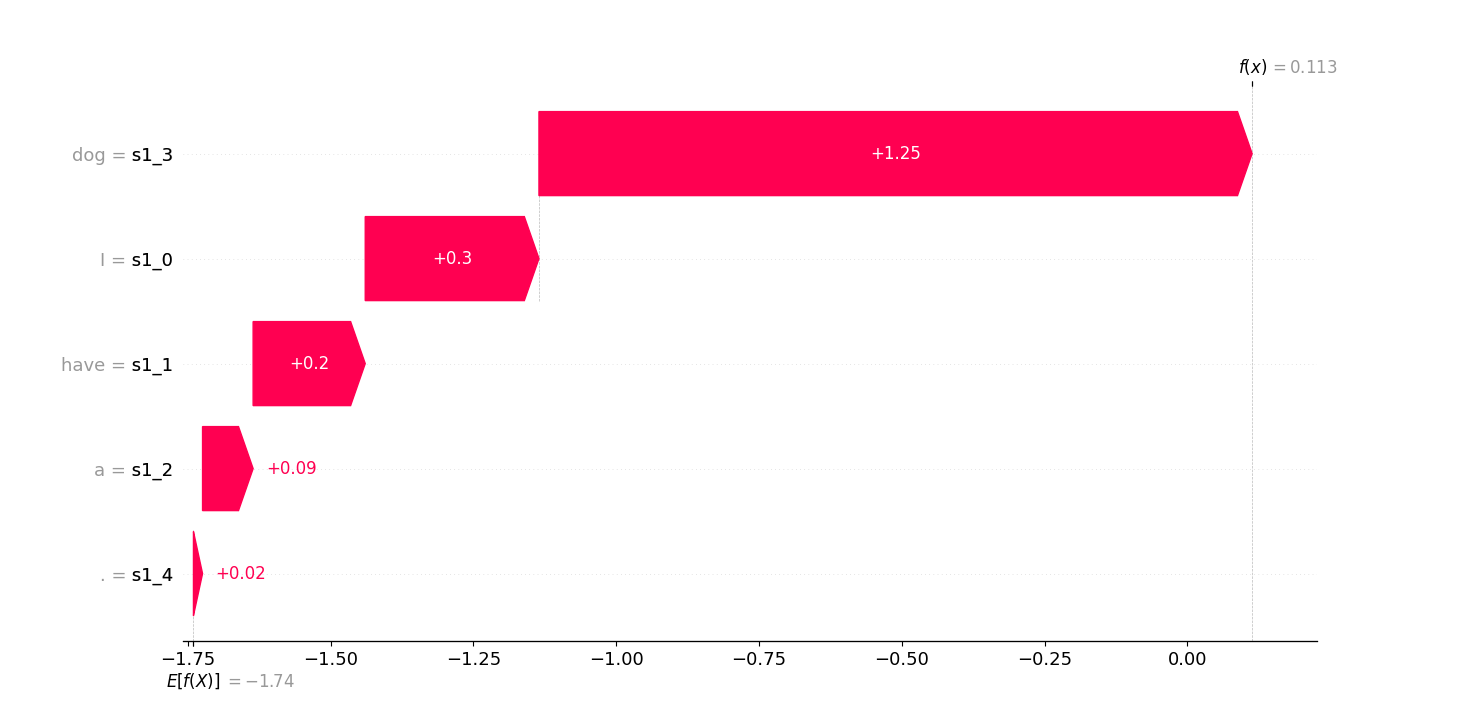}
  \caption{Explanation with respect to a fixed hypothesis}
  \label{fig:hyp_fixed}
\end{figure}

\begin{figure}[]
    \centering
    \includegraphics[width=0.9\textwidth]{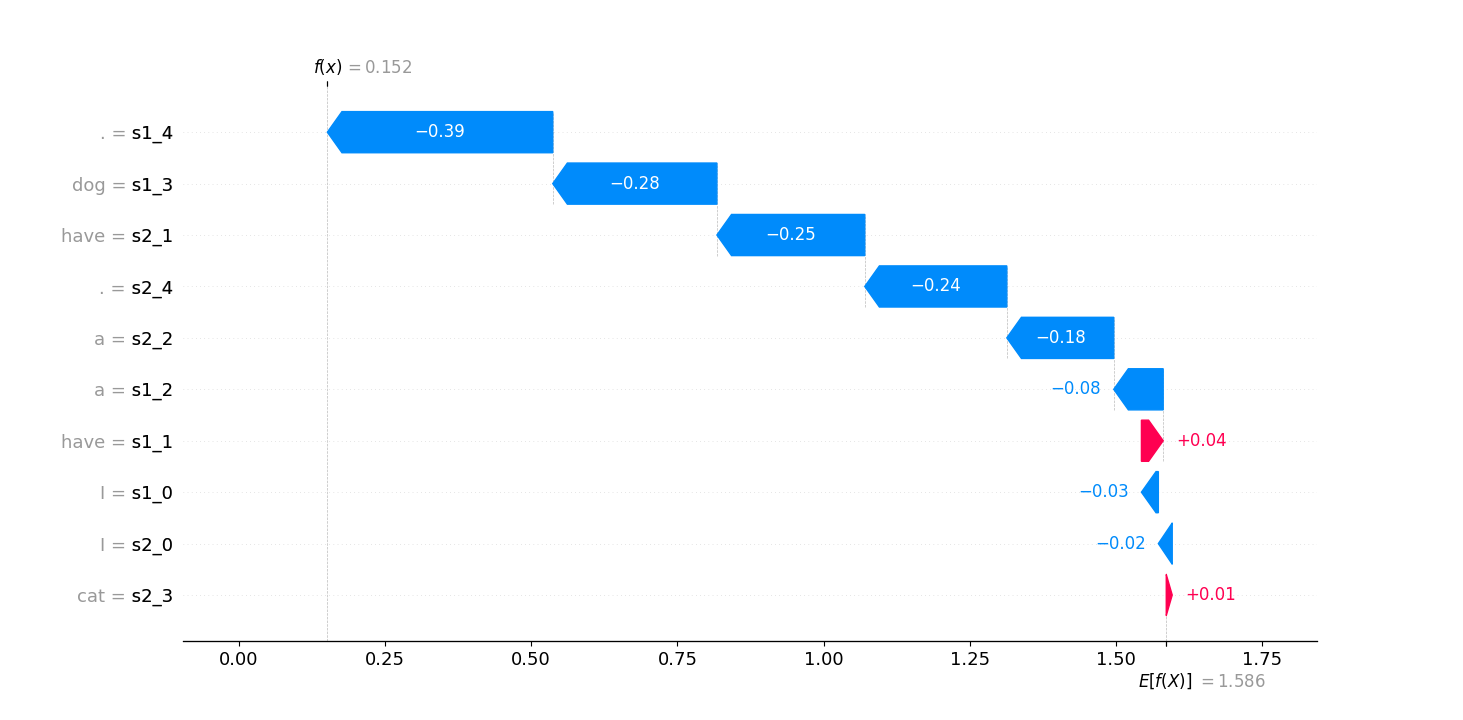}
  \caption{Explanation with free reference and hypothesis}
  \label{fig:free}
\end{figure}

In this section of the Appendix we provide an example that demonstrates the differences in explaining different parts of the input with additive feature attribution methods as described in section \ref{sec:propertiesOfExplainableMTE}. To do so, we extract feature importance scores from BLEURT \citep{sellam-2020} using exact SHAP \citep{lundberg-2017} \footnote{SHAP: \url{https://github.com/slundberg/shap}. BLEURT: \url{https://github.com/google-research/bleurt}}. We mask removed words with an empty string. Further, we explain the score (0.113) that is assigned for the hypothesis ``I have a cat .'' and the reference ``I have a dog .''. In figure \ref{fig:ref_fixed} we fix the reference sentence (i.e. SHAP only perturbs the hypothesis and assigns importance scores to the words in the hypothesis). At the bottom left, we see the Baseline value of -1.859. I.e., BLEURT assigns this score, when the reference sentence is compared to an empty hypothesis. The red arrows indicate how important each of the words is in order to achieve the given score. Compared to its removal, the word cat gets assigned the highest importance, even though it is the only word that was incorrectly translated. This is an interesting inside on the BLEURT model. Further, in \ref{fig:hyp_fixed}, we explain the reference sentence with regards to a fixed hypothesis. Here we similarly see that the word dog was assigned with the highest importance. Finally, in \ref{fig:free}, we explain BLEURT with respect to a free reference and hypothesis. I.e. the importance of all words for achieving a high score in general is evaluated. Here, the baseline is at 1.5 and most words have a negative importance. The reason for this is the similarity of empty sequences, such that the baseline gets highs scores. In general the result is more difficult to interpret.

\end{document}